\documentclass{article}

\usepackage[preprint]{corl_2026}

\usepackage[table]{xcolor}
\usepackage{tikz}
\usetikzlibrary{positioning,fit,backgrounds,decorations.pathreplacing,calc,patterns,arrows.meta,shapes.geometric}
\usepackage[ruled]{algorithm2e}
\usepackage{amsmath}
\usepackage{amssymb}
\usepackage{booktabs}
\usepackage{multirow}
\usepackage{subcaption}
\usepackage{pifont}
\usepackage{hyperref}
\usepackage{enumitem}
\usepackage{tabularx}
\usepackage[capitalise, nameinlink]{cleveref}

\newcommand{\dfa}{\mathcal{D}}
\newcommand{\cmark}{{\color{cgreen}\ding{51}}}
\newcommand{\xmark}{{\color{cred}\ding{55}}}

\newcolumntype{Y}{>{\centering\arraybackslash}X}  

\definecolor{cblue}{HTML}{2563EB}
\definecolor{cred}{HTML}{DC2626}
\definecolor{cgreen}{HTML}{16A34A}
\definecolor{corange}{HTML}{F59E0B}
\definecolor{cgray}{HTML}{6B7280}
\definecolor{clightblue}{HTML}{DBEAFE}
\definecolor{clightgreen}{HTML}{DCFCE7}
\definecolor{clightred}{HTML}{FEE2E2}
\definecolor{clightorange}{HTML}{FEF3C7}
\definecolor{cheaderbg}{HTML}{E5E7EB}    

\title{
Meta-Ctrl: Guaranteed Plan Generation by Decoupling Syntactic and Semantic Constraints 
}

\author{%
  Gwen Yidou-Weng$^{1*}$, Edward Sun$^{1*}$, Tianyi Ma$^{2}$, Metin Alp Dogan$^{1}$,\\
  \textbf{Benjie Wang$^{1}$, Allen Peng$^{1}$, Guy Van den Broeck$^{1}$, Yuchen Cui$^{1}$}\\[4pt]
  $^{1}$University of California, Los Angeles \quad $^{2}$Michigan State University\\
  \texttt{\{gwenweng,edwardsun12895\}@ucla.edu} \quad
  \texttt{\{guyvdb,yuchencui\}@cs.ucla.edu}\\[2pt]
  {\small $^{*}$Equal contribution.}\\
}

\begin{document}
\maketitle

\begin{abstract}
LLMs generate fluent plans for robots but routinely violate the syntactic and
semantic constraints they must satisfy to execute, and existing remedies trade formal guarantees against plan quality: soft methods (affordance scoring, grounded decoding) give no guarantee, while symbolic planners (LLM+P) discard the LM's commonsense. We propose \textbf{Meta-Ctrl}, a constrained-decoding framework that guarantees the encoded constraints while preserving the base LM's plan quality. Meta-Ctrl introduces \emph{meta-tokens}---a compact vocabulary of grounded actions---enforcing syntax at the token level and semantics (preconditions, goals, ordering) at the action level, an exact factorization that cuts the memory of constrained decoding from over $107$\,TB to under $2$\,GB. With it, a small open-weight LM becomes competitive where it otherwise sits at the bottom of the leaderboard: on WAH-NL under the LoTa-Bench protocol it reaches the highest reported subgoal success rate, exceeding GPT-4's, with consistent gains across the Embodied Agent Interface. We further demonstrate it on a real tabletop robot, where every generated plan satisfies its preconditions and goals by construction. 
Project website: \url{https://metactrlg.github.io}
\end{abstract}
\keywords{Constrained Decoding, Formal Guarantees, Task and Motion Planning}

\section{Introduction}

Large language models (LLMs) are increasingly used to generate action plans for
robots, mapping natural-language instructions to sequences of grounded
actions~\cite{ahn2022can, liang2023code, singh2022progprompt, huang2022inner, huang2022language, huang2023grounded}.
On long-horizon household tasks in environments such as
VirtualHome~\cite{puig2018virtualhome} and BEHAVIOR~\cite{srivastava2022behavior},
these plans violate constraints at high rates: hallucinated objects, precondition
failures, missing steps, and incorrect orderings~\cite{li2024embodied, valmeekam2023planbench}.
The failures are of two kinds: \textbf{syntactic} (malformed output: invalid
actions, wrong argument structure, ill-formed formatting) and \textbf{semantic}
(well-formed but physically infeasible or goal-violating, such as placing an
ungrasped object or skipping a required precondition).

Existing methods mitigate these failures but rarely provide both \emph{formal
guarantees} and high plan quality. Affordance scoring~\cite{ahn2022can} and
grounded decoding~\cite{huang2023grounded} reshape the LM's output distribution
with feasibility scores or environment models, but the guidance is soft. Infeasible
outputs can still be generated. Replanning with execution
feedback~\cite{huang2022inner, wang2023voyager} retries after failures, with no
termination guarantee and compounding cost. Active knowledge
acquisition~\cite{liu2026activevoo} learns missing task information by
interaction; it is complementary to enforcing constraints already known, and
could supply the learned constraint extractors our limitations call for.
LLM+P~\cite{liu2023llm+} hands planning
to a PDDL solver, trading the LM's commonsense for a full domain specification.

A separate line of work guarantees validity by construction. Grammar-constrained decoding (GCD)~\cite{poesia2022synchromesh, dong2025xgrammar, willard2023efficient}, the closest to ours, masks at each decoding step any token that would violate the grammar. But the mask is \emph{local} and admits any token
compatible with \emph{some} valid completion, regardless of whether the remaining
sequence is likely or satisfies longer-range constraints, which can sharply
distort the LM's distribution~\cite{park2024grammar}, yielding
valid but improbable outputs, exactly when LM likelihood tracks plan plausibility. Avoiding
this requires \textbf{reasoning about the full remaining sequence, not just the
next token}. Ctrl-G~\cite{zhang2023tractable} does so while keeping the guarantee: it weights each token by a tractable estimate of the probability that the full completion will satisfy the constraint, recovering the LM's distribution conditioned on satisfaction. Applied to syntax, it yields guaranteed well-formed plans directly.

\begin{figure}[t]
    \centering
    \includegraphics[width=\linewidth]{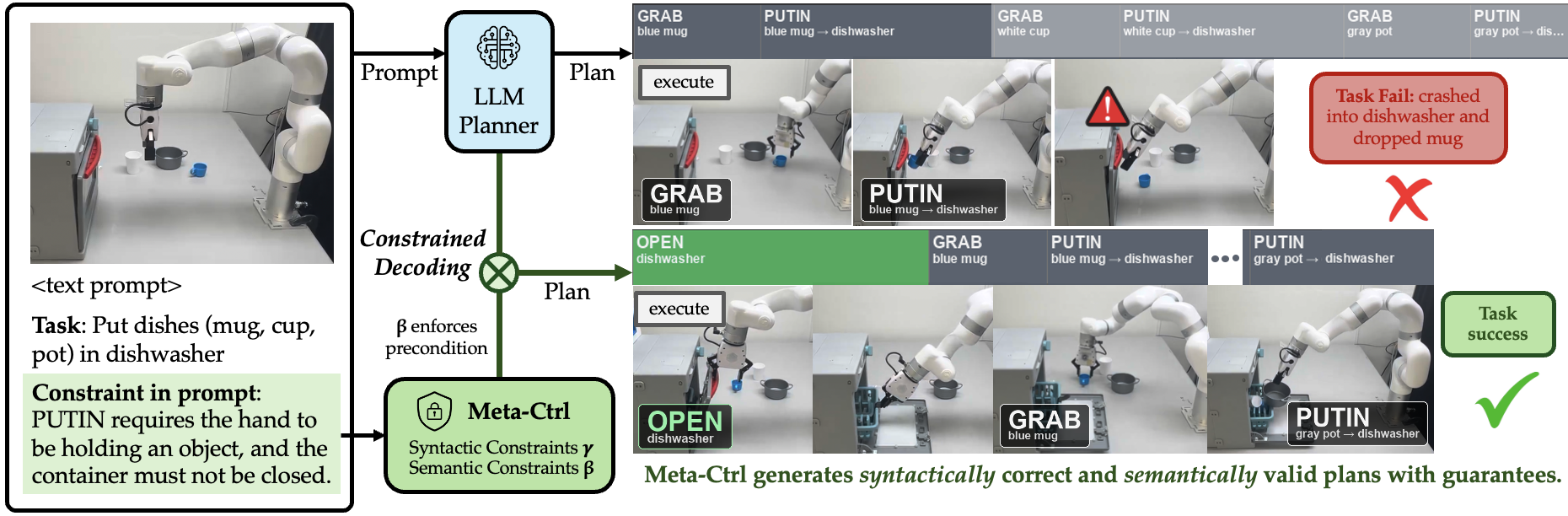}
    \caption{\small{\textbf{Meta-Ctrl enforces preconditions that the base LM violates.}
  Task: put dishes in dishwasher; \textsc{PUTIN} requires holding an object
  \emph{and} an open container. The base LM (top) attempts \textsc{PUTIN} while the
  dishwasher is closed. Meta-Ctrl (bottom) enforces the action-level precondition,
  inserting \textsc{OPEN}(dishwasher).}}
  \vspace{-1.5em}
    \label{fig:teaser}
\end{figure}

Enforcing \emph{semantic} constraints on top of syntax is substantially harder:
semantic validity depends on the evolving world state (e.g.\ whether an object has
been grasped before it is placed), not just each action's surface form. The
na\"{\i}ve fix (compiling syntax and semantics into a single token-level DFA) has
a state space that scales as the \emph{product} of the two, exceeding $350$M states
and requiring $107$\,TB of memory for a typical task. This is wasteful: semantics depends
only on which actions occur and in what order, not on token-level formatting. It
motivates \emph{factoring} constraint enforcement across granularities: syntax at
the token level, semantics at the action level.

We introduce \textbf{Meta-Ctrl}, a two-level constrained-decoding framework built
around \textbf{meta-tokens}: a compact action-level representation that lets syntax
be enforced over tokens and semantics over actions (\cref{fig:teaser}). Because the
two levels can be aligned exactly, the joint constraint decouples into two
independent subproblems, yielding a $1{,}900\times$ reduction in compute and a
$67{,}000\times$ reduction in memory over the monolithic approach. Empirically, this turns a small open-weight LM into a competitive planner that matches or exceeds models an order of magnitude larger (Section~\ref{sec:experiments}).

Our contributions are:
\begin{itemize}[leftmargin=1.2em,itemsep=2pt,topsep=2pt]
\item We bring \textbf{formally guaranteed constrained decoding} to robot planning
  via Ctrl-G, enforcing syntactic constraints with provable validity.
\item We present \textbf{Meta-Ctrl}, a two-level constrained-decoding framework that decouples
token-level syntax from action-level semantics via meta-tokens, an exact factorization of the
joint constraint.
\item We validate Meta-Ctrl on the EAI benchmark~\cite{li2024embodied}
  (VirtualHome and BEHAVIOR), LoTa-Bench~\cite{choilota} (WAH-NL), and a real
  tabletop robot: on VirtualHome action sequencing, an open-weight LM (Llama 3 8B) surpasses every model on the EAI leaderboard, including the
  $8\times$ larger Llama 3 70B and frontier closed models such as GPT-4o,
  Claude 3.5 Sonnet, and o1-preview.
\end{itemize}

\section{Preliminaries: Ctrl-G}
\label{sec:prelim}


\textbf{Constrained decoding as probabilistic reasoning.}
Ctrl-G~\cite{zhang2023tractable} formulates constrained decoding as probabilistic
reasoning over logical constraints. Given an LM and a constraint $\alpha$ (e.g.\
valid JSON, or containing a keyphrase), we sample from
$p_{\mathrm{LM}}(x_{1:n}\mid\alpha)$; decomposing autoregressively,
\begin{equation}
\label{eq:conditional}
  p_{\mathrm{LM}}(x_t \mid x_{<t}, \alpha)
  \;\propto\;
  p_{\mathrm{LM}}(x_t \mid x_{<t})
  \;\cdot\;
  p_{\mathrm{LM}}(\alpha \mid x_{\le t}).
\end{equation}
The first factor is the LM's next-token distribution; the second,
$p_{\mathrm{LM}}(\alpha \mid x_{\le t})$, is the probability that the \emph{full}
sequence will satisfy $\alpha$ given the current prefix. Rather than a hard mask
that treats all locally compatible tokens equally---and so can commit to a token
from which every completion is a dead end---this term \emph{reweights} each token
by the probability that it leads to a constraint-satisfying completion. The
reweighting (i) assigns zero probability to tokens with no valid completion, so
constraint satisfaction is \textbf{guaranteed}, and (ii) among valid tokens,
prefers those leading to natural, high-probability completions under the LM,
preserving fluency.

\textbf{Tractable lookahead via HMM and DFA.}
Computing $p_{\mathrm{LM}}(\alpha \mid x_{\le t})$ exactly is intractable, as it
marginalizes over all future sequences that satisfy $\alpha$. Ctrl-G trains a
Hidden Markov Model $p_{\mathrm{HMM}}(x_{1:n}) \approx p_{\mathrm{LM}}(x_{1:n})$ and
substitutes $p_{\mathrm{HMM}}(\alpha \mid x_{\le t})$:
\begin{equation}
\label{eq:ctrlg}
  p_{\mathrm{ctrl}}(x_t \mid x_{<t}, \alpha)
  \;\propto\;
  p_{\mathrm{LM}}(x_t \mid x_{<t})
  \;\cdot\;
  p_{\mathrm{HMM}}(\alpha \mid x_{\le t}).
\end{equation}
The guarantee is preserved---$p_{\mathrm{HMM}}(\alpha \mid x_{\le t})=0$ wherever no
continuation satisfies $\alpha$---so the HMM affects only the reweighting among
valid tokens, and a more accurate HMM yields generations that are both
constraint-satisfying and natural. To evaluate the lookahead, $\alpha$ is
represented as a deterministic finite automaton (DFA) $\dfa_\alpha$ that reads
tokens left to right and accepts in a final state: a natural match for
autoregressive decoding, composable by intersection and union, with size $|\dfa|$
measured in edges. The HMM and DFA are both finite-state, so their product is a
finite-state model over (hidden state, DFA state) pairs, admitting a backward
dynamic program (DP) that returns $p_{\mathrm{HMM}}(\alpha \mid x_{\le t})$ for
every prefix; for length $n$, $H$ hidden states, and DFA size $|\dfa|$, it costs
$O(n \cdot |\dfa| \cdot H^2)$. Ctrl-G was demonstrated on constraints with small
DFAs (keywords, length, infilling); for compositional constraints the standard
product of DFAs has edge count $|\dfa_1|\cdot|\dfa_2|$, so the DP cost scales
multiplicatively---the bottleneck we address next.

\section{Decoupling Syntax and Semantics for Constrained Decoding}
\label{sec:method}

Embodied plans must satisfy two kinds of constraints at different granularities:
a token-level \textbf{syntactic} constraint $\gamma$ (valid action names, argument
structure, well-formed JSON) and an action-level \textbf{semantic} constraint
$\beta$ (goal achievement, preconditions, temporal ordering among grounded
actions).
We enforce each at its natural level,
applying Ctrl-G first to syntax (Section~\ref{sec:level1}),
then extending to semantics through a meta-token projection
(Section~\ref{sec:level2}),
and finally showing why factoring across the two levels
is both necessary and beneficial (Section~\ref{sec:whytwolevels}).

\subsection{Level 1: Token-Level Syntax}
\label{sec:level1}

The syntactic constraint is naturally a token-level DFA $\dfa_\gamma$
that accepts well-formed action blocks:
valid action names, correct argument arity and types,
and proper JSON structure.
Applying Ctrl-G with $\dfa_\gamma$ steers generation toward
syntactically valid output:
\begin{equation}
\label{eq:level1}
  p(x_t \mid x_{<t}, \gamma)
  \;\propto\;
  p_{\mathrm{LM}}(x_t \mid x_{<t})
  \;\cdot\;
  p_{\mathrm{HMM}}(\gamma \mid x_{\le t}).
\end{equation}
This eliminates malformed output entirely: the raw LM frequently emits prose
(e.g.\ \texttt{``To solve this, the action is TYPE\ldots''}), invalid action
names, or ill-formed arguments, whereas every Level-1 output parses into a valid
action sequence (e.g.\ \texttt{\{"WALK":["computer"],"GRAB":["mouse"]\}}).

\textbf{Syntax is not enough.}
A syntactically valid plan can still be \emph{semantically} wrong:
it may attempt to place an object before grasping it,
open a container that is already open,
or terminate before achieving the goal.
These errors naturally live at the \emph{action} level,
not the token level.
We therefore need a second constraint
that reasons about the action sequence as a whole.

\subsection{Level 2: Action-Level Semantics via Meta-Tokens}
\label{sec:level2}

\newcommand{\mvocab}{\mathcal{W}}

\begin{figure}[t]
\centering
\begin{tikzpicture}[
  state/.style={ellipse, draw, fill=clightblue!40, minimum width=1.6cm,
                minimum height=0.8cm, font=\scriptsize, inner sep=2pt, align=center},
  accept/.style={state, fill=clightorange!70, double},
  tok/.style={draw, rounded corners=1pt, fill=clightblue!20,
              minimum width=0.5cm, minimum height=0.42cm,
              font=\tiny\ttfamily, inner sep=1pt},
  metatok/.style={draw, rounded corners=2pt, fill=clightgreen!50,
                  minimum width=1.8cm, minimum height=0.75cm,
                  font=\scriptsize, align=center, inner sep=2pt},
]
  \node[font=\scriptsize, anchor=east] at (-0.2, 1.7) {tokens};
  \foreach \i/\t in {0/{[}, 1/\{, 2/", 3/G, 4/RAB, 5/", 6/:, 7/plate, 8/\}, 9/{,}} {
    \node[tok] at (\i*0.6, 1.7) {\t};
  }
  \node[font=\tiny, cgray, anchor=west] at (9*0.6+0.4, 1.7) {\ldots};

  \draw[->, very thick, cgreen!70!black] (1.2, 1.35) -- (1.2, 0.55);
  \draw[->, very thick, cgreen!70!black] (4.2, 1.35) -- (4.2, 0.55);
  \node[font=\scriptsize, anchor=west, cgreen!60!black] at (1.45, 0.95) {project via $\tau$};

  \node[font=\scriptsize, anchor=east] at (-0.2, 0.2) {actions};
  \node[metatok] (m1) at (1.2, 0.2) {\textsc{Grab}\\plate};
  \node[metatok] (m2) at (4.2, 0.2) {\textsc{Open}\\dishwasher};
  \node[metatok] (m3) at (7.0, 0.2) {\textsc{Put}\\plate, dish.};

  \node[font=\scriptsize, anchor=east] at (-0.2, -1.4) {$\beta$ DFA};
  \node[state]  (qC) at (1.2, -1.4) {closed};
  \node[state]  (qO) at (4.2, -1.4) {open};
  \node[accept] (qA) at (7.2, -1.4) {goal};
  \draw[->, thick, gray!70] (qC) to[bend left=22]
    node[above, font=\scriptsize, black] {\textsc{Open}} (qO);
  \draw[->, thick, gray!70] (qO) to[bend left=22]
    node[below, font=\scriptsize, black] {\textsc{Close}} (qC);
  \draw[->, thick, gray!70] (qO) --
    node[above, font=\scriptsize, black, align=center] {\textsc{Put} plate in} (qA);
  \draw[->, very thick, cblue!70!black] (4.2, -0.2) -- (4.2, -0.95);
\end{tikzpicture}
\caption{%
  \small{\textbf{Level 2: action-level semantics.}
  The parser $\tau$ projects each action block of tokens
  (e.g.\ \texttt{\{"GRAB":"plate"\}}) to one meta-token (\textsc{Grab}(plate)).
  A DFA $\dfa_\beta$ over meta-tokens enforces preconditions and goal progress
  (e.g.\ \textsc{Put} requires the container open).
  Operating on a small action vocabulary, rather than raw tokens,
  is what makes the factored computation below tractable.}}
\label{fig:token-to-meta}
\end{figure}
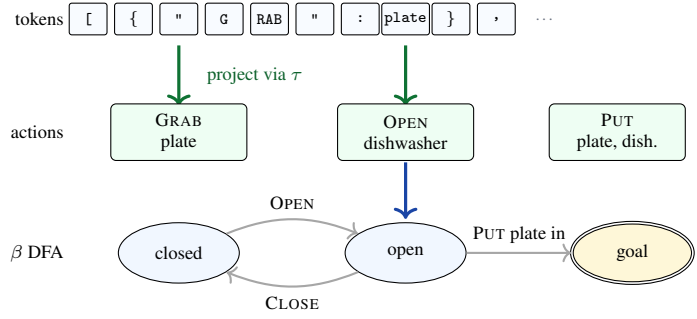

To enforce semantics, we lift generation to the action level.
We introduce a vocabulary $\mathcal{W}$ of \textbf{meta-tokens},
each a grounded action (e.g., \textsc{Open}(dishwasher)),
and a \textbf{parser}
$\tau : \mathrm{supp}_\gamma \to \mathcal{W}^*$
that maps each syntactically valid token sequence
to its action sequence,
much as a tokenizer maps characters to tokens
(Figure~\ref{fig:token-to-meta}).
The semantic constraint $\beta$ is then a DFA \emph{over meta-tokens}
that tracks task-relevant state
(which actions have occurred, whether preconditions hold,
whether the goal is reached).
We assume $\tau$ is a bijection on $\mathrm{supp}_\gamma$,
so every action sequence has a unique token realization
$\tau^{-1}(w)$.\footnote{
For each action sequence to have exactly one token sequence, we fix a deterministic \emph{canonical} tokenizer (typically shipped with the LM) and let $\gamma$ admit only canonically-tokenized sequences. Then $\tau$ is one-to-one.
%
  }
The joint constraint is
\begin{equation}
\label{eq:alpha-def}
  x \models \alpha
  \;\iff\;
  (x \models \gamma) \;\wedge\; (\tau(x) \models \beta).
\end{equation}

\textbf{Factored posterior.}
At decoding step $t$, let $a_{<l} = \tau_{\mathrm{inc}}(x_{\le t})$
be the completed action history,
where $\tau_{\mathrm{inc}}$ returns the actions parsed so far.
Marginalizing the joint posterior over the identity of the current action,
\begin{equation}
\label{eq:factored}
  p(\alpha \mid x_{\le t})
  = \sum_{a_l}
    \underbrace{p(a_l \mid x_{\le t})}_{\text{bridging}}
    \cdot
    \underbrace{p(\gamma \mid x_{\le t}, a_l)}_{\text{syntax}}
    \cdot
    \underbrace{p(\beta \mid x_{\le t}, a_l, \gamma)}_{\text{semantics}}.
\end{equation}
The semantics term still couples both granularities.
We decouple it using the bijectivity of $\tau$ on $\mathrm{supp}_\gamma$.
Define the induced meta-token distribution
\begin{equation}
\label{eq:pmeta-def}
  p_{\mathrm{meta}}(w) \;:=\; p_{\mathrm{LM}}\!\left(\tau^{-1}(w) \,\big|\, \gamma\right),
\end{equation}
which is well-defined because $\tau^{-1}(w)$ is a unique token sequence
in $\mathrm{supp}_\gamma$ for each $w \in \mvocab^*$.
Conditional on the completed actions $a_{<l}$, the candidate next
action $a_l$, and the syntactic constraint $\gamma$,
the token-level prefix $x_{\le t}$ carries no further information about
the meta-token sequence: $a_{<l}$ together with $a_l$ fixes the prefix
of meta-tokens, which by bijectivity fixes the prefix of tokens up to
the canonical tokenization that $\gamma$ enforces.
Hence
\begin{equation}
\label{eq:beta-indep}
  p\!\left(\beta \,\big|\, x_{\le t}, a_l, \gamma\right)
  \;=\; p_{\mathrm{meta}}\!\left(\beta \,\big|\, a_{\le l}\right),
\end{equation}
and substituting into Eq.~\ref{eq:factored} yields the fully factored
posterior
\begin{equation}
\label{eq:final}
  p(\alpha \mid x_{\le t})
  = \sum_{a_l}
    \underbrace{p(a_l \mid x_{\le t})}_{\text{bridging}}
    \cdot
    \underbrace{p(\gamma \mid x_{\le t}, a_l)}_{\substack{\text{syntax}\\(\text{token-level})}}
    \cdot
    \underbrace{p_{\mathrm{meta}}(\beta \mid a_{\le l})}_{\substack{\text{semantics}\\(\text{action-level})}}.
\end{equation}
Syntax is enforced purely at the token level
and semantics purely at the action level,
communicating through the bridging term
$p(a_l \mid x_{\le t})$,
which maps the current partial token sequence
to a distribution over the action it commits to.
The decoupling is exact and requires no assumption on the LM
beyond bijectivity of $\tau$.
Algorithm~\ref{alg:two-level} (Appendix~\ref{apx:algorithm}) summarizes the
resulting two-level decoder.

\subsection{Why Two Levels?}
\label{sec:whytwolevels}

Enforcing $\beta$ and $\gamma$ jointly in a single DFA
would require lifting $\beta$ to the token level
and intersecting with $\gamma$,
producing a joint DFA whose state space is the product
of the two (Appendix~\ref{apx:joint_dfa_construction}).
Factoring keeps the levels separate: it replaces that multiplicative state space with an additive one, advances the semantic DP at the action rather than token granularity, and lets the meta-token HMM concentrate on a small action vocabulary instead of the syntactic variation $\gamma$ already handles. The result is both cheaper and a better surrogate (Table~\ref{tab:whytwolevels}).
This yields a lower per-action NLL surrogate,
and a more accurate surrogate produces higher-quality plans
among the constraint-satisfying set (Section~\ref{sec:prelim}).
The monolithic cost is what exposing the full world state to the token level
requires: the semantic automaton's states are assignments to the scene's
tracked object attributes, so its size is exponential in scene detail, and
intersecting it with the token-level DFA multiplies the two, giving the
$350$M-state, $107$\,TB object above. Grammar-constrained decoders never meet
this cost because they represent no world state, which is equally why they
enforce syntax only; enforcing semantics requires carrying the state, and
factoring is what makes carrying it affordable.
The factorization also separates the pipeline's cheap and expensive parts: each task's
constraints live in DFAs that compile in seconds and swap freely, while the costly trained HMMs
are few and shared, one meta-token HMM per module type and one token-level HMM per base LM
(Appendix~\ref{apx:hmm}).

\begin{table}[t]
\caption{%
\small{  Why two levels: factoring the joint constraint
  is both cheaper and produces a better surrogate.
  Numbers are for a median VirtualHome task ($H{=}128$); see Section~\ref{sec:experiments} for the quality ablation. Both $-\log p$ columns are per grounded action; sequence length is raw tokens
(42.7) for the token HMM and grounded actions (7.6) for the meta-token HMM.
  }
  }
\label{tab:whytwolevels}
\centering
\small
\begin{minipage}{0.48\textwidth}
\centering
\textbf{Efficiency}\\[0.3em]
\begin{tabular}{lrr}
\toprule
& Monolithic & Two-level \\
\midrule
States        & $S_\gamma{\times}S_\beta$ & $S_\gamma{+}S_\beta$ \\
              & 350\,M & 57\,K \\
Compute       & 1{,}720\,T & 922\,G \\
DP memory     & 107\,TB & 1.6\,GB \\
\midrule
\multicolumn{3}{c}{${\sim}1{,}900\times$ faster, ${\sim}67{,}000\times$ less memory} \\
\bottomrule
\end{tabular}
\end{minipage}
\hfill
\begin{minipage}{0.48\textwidth}
\centering
\textbf{Quality}\\[0.3em]
\begin{tabular}{lcc}
\toprule
& Token HMM & Meta-token HMM \\
\midrule
Vocab          & 128\,K & 132 \\
Emission       & 16.4\,M & 16.9\,K \\
Seq.\ length   & 42.7 & 7.6 \\
$-\log p$/act.  & 33.54 & \textbf{2.01} \\
\midrule
\multicolumn{3}{c}{${\sim}16.7\times$ better fit} \\
\bottomrule
\end{tabular}
\end{minipage}
\end{table}

\section{Experiments}
\label{sec:experiments}

We evaluate Meta-Ctrl on three planning benchmarks
and on a real XArm7
tabletop robot (Section~\ref{sec:exp-robot}). A lookahead ablation isolates probabilistic
reweighting as the source of the gains, and an analysis
(Appendix~\ref{sec:dfa-info-audit}) explains where Meta-Ctrl helps most.

\subsection{Setup}
\label{sec:setup}
\textbf{Benchmarks.}
We evaluate on three benchmarks. \emph{Embodied Agent Interface}
(EAI)~\cite{li2024embodied} covers VirtualHome (VH) and BEHAVIOR (BEH)
over two plan-generation modules, \emph{Action Sequencing} (AS) and
\emph{Subgoal Decomposition} (SD), yielding four (simulator, module)
sets of $342$, $338$, $100$, and $100$ tasks.\footnote{The other two
EAI modules (\emph{Goal Interpretation}, \emph{Transition Modeling})
are structured-prediction tasks whose failure modes are missing or
misinterpreted predicates, which constraint enforcement cannot supply.}
\emph{Watch-And-Help} (WAH-NL)~\cite{puigwatch}, under the LoTa-Bench
protocol~\cite{choilota}, provides $100$ single-agent tasks across five
families with a stricter symbolic executor than EAI. Each task provides
an instruction and a structured scene description and is scored by the
official evaluator on goal satisfaction and executability. We also
deploy on a \emph{real XArm7 tabletop} with open-vocabulary perception
and downstream execution modules (Section~\ref{sec:exp-robot}).

\textbf{Base LMs and baselines.}
On EAI, we evaluate Meta-Ctrl on two open-weight base models,
\texttt{Llama-3-8B-Instruct}~\cite{grattafiori2024llama} and
\texttt{gpt-oss-20B}~\cite{openai2025gptoss120bgptoss20bmodel}; on
WAH-NL, we use \texttt{Llama-3.1-8B-Instruct}~\cite{grattafiori2024llama}.
We compare against all 14 EAI leaderboard models~\cite{li2024embodied}
and, on WAH-NL, published numbers for SayCan~\cite{ahn2022can},
ProgPrompt~\cite{singh2022progprompt}, LoTa-Bench~\cite{choilota}, and
STEP~\cite{zhou2025step}. For internal comparison we also report each
base LM unconstrained and under a hard-masking variant of the syntactic
DFA.

\textbf{Metrics.} We report \emph{Task SR} (fraction of plans achieving
the goal) and \emph{Exec SR} (fraction executing to completion in the
simulator); WAH-NL additionally reports \emph{SSR}, the fraction of
subgoals satisfied. Real-robot metrics are defined in
Section~\ref{sec:exp-robot}.

\textbf{Meta-Ctrl configuration.} We build a syntactic $D_\gamma$ and a
semantic $D_\beta$ per (benchmark, module) pair and induce $\tau$ from
$D_\gamma$. The trained HMMs are shared across pairs rather than built
per task; Appendix~\ref{apx:hmm} gives the sharing structure and
training details. All train on unconstrained base-LM generations
matching the benchmark format but containing no evaluation instance.
Decoding is greedy under Eq.~\ref{eq:final} with a 40-action horizon
for the semantic DP; provably infeasible tasks fall back to
$\gamma$-only.
 

\subsection{Watch-And-Help via LoTa-Bench}
\label{sec:exp-wah}
\begin{table}[t]
\caption{%
\small{
  WAH-NL (single-agent, $n=100$) under the LoTa-Bench protocol.
  External references use different LMs and executors; 
  we report them as ballpark context, not strict baselines. Empty entries are not reported.
  Same-LM rows (Llama 3.1 8B) are direct comparisons.
  $^\dagger$STEP is closed-loop; ours is open-loop.}
  }
\label{tab:wah}
\centering
\small
\begin{tabularx}{\linewidth}{lXYYY}
\toprule
\textbf{Method} & \textbf{Base LM} & \textbf{SR} & \textbf{SSR} & \textbf{Exec} \\
\midrule
\multicolumn{5}{l}{\textit{External references (different LM / executor)}} \\
\quad SayCan                    & ---              & 0.010 & 0.021 & --- \\
\quad ProgPrompt                & ---              & 0.030 & 0.187 & --- \\
\quad LoTa-Bench                & GPT-4            & ---   & 0.342 & --- \\
\quad LoTa-Bench                & LLaMA-1 65B    & ---   & 0.433 & --- \\
\quad STEP$^\dagger$            & (larger)         & 0.400 & 0.620 & --- \\
\midrule
\multicolumn{5}{l}{\textit{Same LM (Llama 3.1 8B), same evaluator, $n=100$}} \\
\quad Raw LM (unconstrained)    & Llama 3.1 8B     & 0.000 & 0.022 & 0.010 \\
\quad + syntax       & Llama 3.1 8B     & 0.000 & 0.037 & 0.010 \\
\quad \textbf{Meta-Ctrl} & Llama 3.1 8B     & \textbf{0.470} & \textbf{0.705} & \textbf{1.000} \\
\bottomrule
\end{tabularx}
\end{table}

\textbf{Constrained decoding turns an unusable LM into a deployable one, and
semantic control drives the gain.} \texttt{Llama-3.1-8B} alone solves none of the $100$ tasks
(SR $0.000$, Exec $0.010$); its raw outputs are already mostly well-formed at the
token level, so syntactic control alone ($+\gamma$) raises SSR by just $1.5$ points
and leaves Exec at $0.010$. Adding semantic control (full Meta-Ctrl) lifts SSR a
further $66.8$ points and Exec by $99$, reaching SR $0.470$, SSR $0.705$,
Exec $1.000$. On the same LM and evaluator, task success goes from $0$ to
$0.47$, and semantic control accounts for nearly all of it. On the SSR axis comparable
across protocols, Meta-Ctrl ($0.705$) exceeds LoTa-Bench's strongest reported
configuration (LLaMA-1-65B, $0.433$; GPT-4, $0.342$) and the closed-loop STEP
baseline ($0.620$).\footnote{Prior WAH-NL evaluations use natural-language
instruction inputs and a Unity simulator; the LoTa-Bench protocol~\cite{choilota}
uses oracle-goal inputs and a strict symbolic executor. SSR-to-SSR is the
comparable axis across protocols.}

\textbf{The constraint enforces the invariants that prompting baselines
violate.} Examining where the ProgPrompt baseline's plans fail under the
same executor: $60$ of $100$ fail at \textsc{Grab}, $21$ at \textsc{PutIn},
$10$ at \textsc{PutBack}. The failures are precondition violations (hands
already full, target not co-located, container closed), not formatting or
hallucination. Meta-Ctrl's $\beta$ DFA tracks these invariants at
every decoding step, and enforcing them during generation, instead of
checking them after the plan is produced, is what yields the $0\to 1.000$
execution rate.

\subsection{Embodied Agent Interface}
\label{sec:exp-eai}

\begin{table}[t]
\caption{%
\small{
  Embodied Agent Interface results on VirtualHome (V; $n{=}342$ AS, $338$ SD)
  and BEHAVIOR (B; $n{=}100$ AS, $100$ SD).
  Task SR and Execution SR (\%). Leaderboard baselines from~\cite{li2024embodied};
  Each base LM is grouped with its full Meta-Ctrl ($\gamma{+}\beta$) result.
  \textbf{Bold}: best per column; \underline{underline}: second-best.
  $^{\ddagger}$The gpt-oss rows use a keyword-based BEHAVIOR affordance
  prior; the Llama rows use the taxonomy-derived specification of
  Appendix~\ref{app:beh-affordance}.
  Our VirtualHome cells are means over six evaluator seeds
  (Appendix~\ref{apx:evalnoise}).
  }
  }
\label{tab:eai-main}
\centering
\small

\begin{tabularx}{\linewidth}{lYYYYYYYY}
\toprule
& \multicolumn{4}{c}{\textbf{Action Sequencing}}
& \multicolumn{4}{c}{\textbf{Subgoal Decomposition}} \\
\cmidrule(lr){2-5}\cmidrule(lr){6-9}
& \multicolumn{2}{c}{Task SR} & \multicolumn{2}{c}{Exec SR}
& \multicolumn{2}{c}{Task SR} & \multicolumn{2}{c}{Exec SR} \\
\textbf{Model} & VH & BEH & VH & BEH & VH & BEH & VH & BEH \\
\midrule
\rowcolor{cheaderbg}
\multicolumn{9}{l}{\textit{Representative baselines}} \\
GPT-4o                                  & 71.5 & 47.0 & 81.3 & 53.0 & 87.6 & \underline{49.0} & 91.1 & 55.0 \\
Claude-3.5 Sonnet                       & 76.1 & \underline{60.0} & 81.3 & 69.0 & 89.1 & 39.0 & 92.0 & 44.0 \\
o1-preview                              & 65.2 & \textbf{81.0} & 72.5 & \textbf{91.0} & \underline{89.4} & \textbf{60.0} & \underline{93.2} & 62.0 \\
Mistral Large                           & 78.4 & 33.0 & 84.6 & 50.0 & 84.3 & 31.0 & 92.0 & 38.0 \\
Llama 3 70B Instruct                    & 59.0 & 34.0 & 66.6 & 42.0 & 78.4 & 21.0 & 87.3 & 30.0 \\
\midrule
Llama 3 8B Instruct (base)              & 21.3 & 10.0 & 23.6 & 16.0 & 48.8 & 22.0 & 58.0 & 29.0 \\
\quad + Meta-Ctrl (ours)                & \textbf{90.4} & 34.0 & \textbf{97.7} & 89.0 & \textbf{89.9} & 36.0 & \textbf{93.5} & \textbf{66.0} \\
\midrule
gpt-oss-20B (base)                      & 74.4 & 40.0 & 80.3 & 51.0 & 72.5 & 27.0 & 82.2 & 36.0 \\
\quad + Meta-Ctrl (ours)$^{\ddagger}$   & \underline{86.6} & 40.0 & \underline{94.1} & \underline{90.0} &  82.3  & 41.0 &  86.4  & \textbf{66.0} \\
\bottomrule
\end{tabularx}
\end{table}

\textbf{Meta-Ctrl lifts open-weight LMs to match frontier closed-weight
models on VirtualHome.} On \texttt{Llama-3-8B-Instruct}, Meta-Ctrl raises task SR
from $21.3$ to $90.4$ on VH Action Sequencing and from $48.8$ to $89.9$
on Subgoal Decomposition (Table~\ref{tab:eai-main}); on \texttt{gpt-oss-20B}, it
raises VH AS from $74.4$ to $86.6$. Both lifted numbers exceed every
model on the EAI leaderboard, including o1-preview ($65.2$ on VH AS),
Claude 3.5 Sonnet ($76.1$), and Mistral Large ($78.4$). On BEHAVIOR the
task-SR lift is positive but smaller (\texttt{Llama-3-8B}: $10\to 34$ on AS,
$22\to 36$ on SD; \texttt{gpt-oss}: $40\to 40$ on AS, $27\to 41$ on SD), while
the execution-SR lift stays large ($16\to 89$ on AS, $29\to 66$ on SD):
enforcing the constraints removes the failures that stop a plan from running,
but achieving a BEHAVIOR goal also needs world knowledge the prompt withholds
(Appendix~\ref{sec:dfa-info-audit}). Constrained
decoding closes the gap between an $8$B open-weight LM and the strongest
proprietary models on VirtualHome and brings a $20$B open-weight LM
above all of them.

\textbf{Meta-Ctrl works best when the prompt specifies the task's
constraints.} The gap between VirtualHome and BEHAVIOR shows when
Meta-Ctrl helps most. VirtualHome prompts describe each object's
properties and states directly, so the DFA can encode $82\%$ of the
task's preconditions from the prompt alone; BEHAVIOR prompts
do not, and the DFA can encode only $24\%$ from the prompt
(Appendix~\ref{sec:dfa-info-audit}). For BEHAVIOR the remainder comes from the
benchmark's public BDDL object taxonomy, looked up by the synsets the prompt
already names, plus five documented generic rules
(Appendix~\ref{app:beh-affordance}); this is still strictly less than the complete
transition model the benchmark hands its own planner baselines. The method
recovers the most when a task's constraints can be made explicit and written
into the DFA.

\begin{table}[t]
\caption{%
  \small{Porting Meta-Ctrl to newer base LMs, VH AS ($n{=}342$), task /
  execution SR (\%). Each port rebuilds $\gamma$ over the new tokenizer
  automatically and uses a uniform token-level HMM over the new vocabulary;
  the meta level is unchanged (uniform meta-surrogate rows for all three
  bases). No training of any kind. VH-AS means over six evaluator seeds.}}
\label{tab:newbases}
\centering
\small
\begin{tabularx}{\linewidth}{lYYY}
\toprule
\textbf{Base LM} & \textbf{unconstrained} & \textbf{+ Meta-Ctrl} & \textbf{changed lines} \\
\midrule
Llama 3 8B Instruct   & 21.3~/~23.6 & 90.7~/~97.7 & --- \\
Qwen3-8B              & 50.8~/~58.4 & 90.7~/~97.7 & ${\sim}40$ \\
Gemma-3-12B-it        & 47.5~/~50.5 & 89.4~/~96.1 & ${\sim}35$ \\
\bottomrule
\end{tabularx}
\end{table}

\textbf{The construction transfers to new base LMs without training.}
The token-level HMM turns out to be nearly inert once $\beta$ is present:
replacing the trained one with a uniform HMM over the same vocabulary leaves
$273$ of $342$ VH AS programs byte-identical, and the $69$ that differ swap
tie-broken synonym verbs ($41$ of them \textsc{Find}$\leftrightarrow$\textsc{Walk}),
moving task SR by $-0.7$. Since $\gamma$ is compiled from the prompt over
whatever tokenizer the base model uses, and the meta level is defined over
grounded actions rather than tokens, porting to a new base LM needs no
retraining at either level. We ported to two 2025 open-weight models spanning
two tokenizer families, Qwen3-8B (byte-level BPE, $152$K) and
Gemma-3-12B-it (SentencePiece, $262$K), in well under $100$ changed lines
each, and both reproduce the pattern: both task and execution SR transfer exactly on
Qwen3 ($90.7$/$97.7$, tying the Llama pipeline), and the guarantee transfers
with zero parsing and zero affordance errors on both (\cref{tab:newbases}).

\begin{table}[t]
\caption{%
  \small{
  What each component contributes, task / execution SR (\%) on Llama 3 8B.
  Rows 3--6 share the \emph{same} constraint construction (meta-tokens,
  $\gamma$ and $\beta$ compiled from the prompt and the public object
  taxonomy) and the same guarantee; they differ only in the guidance signal.
  \emph{Search, no LLM} is breadth-first search over the meta-level automaton
  with the LM removed. \emph{Masking} keeps only the support of the lookahead
  term, a reachability yes/no. \emph{Uniform} and \emph{trained} instantiate
  the lookahead measure itself. \emph{Classical planner} is Fast Downward
  given the benchmark's own PDDL domain, i.e.\ a complete transition model,
  strictly more information than our automata encode; SD asks for goal states
  rather than actions, so planners do not apply.
  }}
\label{tab:lookahead-ablation}
\centering
\small
\begin{tabularx}{\linewidth}{lYYYY}
\toprule
\textbf{Method (Llama 3 8B)} & \textbf{VH AS} & \textbf{VH SD} & \textbf{BEH AS} & \textbf{BEH SD} \\
\midrule
LLM only (unconstrained)             & 21.3~/~23.6 & 48.8~/~58.0 & 10~/~16 & 22~/~29 \\
LLM + syntax-only GCD ($\gamma$ mask) & ~1.4~/~97.3 & ~0.0~/~96.0 & 12~/~20 & 19~/~31 \\
Classical planner, full domain       & 47.2~/~84.6 & n/a         & 13~/~37 & n/a \\
\midrule
\multicolumn{5}{l}{\emph{with our meta-level automata:}}\\
Search, no LLM                       & 83.8~/~93.1 & 72.5~/~79.0 & 31~/~77 & 52~/~64 \\
LLM + masking (reachability only)    & 81.7~/~87.5 & 88.7~/~92.6 & 33~/~85 & 38~/~51 \\
LLM + uniform lookahead              & \textbf{90.7}~/~\textbf{97.7} & 88.4~/~93.2 & \textbf{36}~/~89 & \textbf{41}~/~59 \\
LLM + trained lookahead (Meta-Ctrl)  & 90.4~/~\textbf{97.7} & \textbf{89.9}~/~\textbf{93.5} & 34~/~89 & 36~/~\textbf{66} \\
\bottomrule
\end{tabularx}
\end{table}

\textbf{Guarantees.} Format and hallucination errors that account for
$24\%$ of raw \texttt{Llama-3-8B} outputs on BEH AS ($15\%$ hallucinated objects,
$9\%$ malformed parameters) drop to $0\%$ under Meta-Ctrl by
construction; the same pattern holds on VH AS, where the LM's $22.3\%$
combined error rate becomes $0\%$ LM-fault under Meta-Ctrl
(Appendix~\ref{apx:format-errors} reports the per-module breakdown).
The guarantee eliminates the failure modes that account for the majority
of grammar errors at this scale.

\textbf{Semantic constraints and probabilistic lookahead each carry part of
the gain.} Table~\ref{tab:lookahead-ablation} separates them. Syntax-only
masking, the standard grammar-constrained setup, collapses task success while
inflating execution ($1.4$ task at $97.3$ execution on VH AS): the greedy LM
steers toward the shortest well-formed continuation, which executes cleanly
and achieves nothing. The decoding target is
$p(a \mid \text{constraint satisfiable}) \propto p_{LM}(a)\cdot
P(\text{satisfiable future} \mid a)$, and the second factor is a probability;
masking, uniform, and trained lookaheads are successive approximations of it.
Masking keeps only its support, a reachability yes/no that tells the LM
nothing about the future: under identical $\gamma{+}\beta$ automata it reaches
$81.7$ on VH AS, no better than searching the automaton with no LM at all.
The uniform lookahead scores each step by the accepting future it preserves
and reaches $90.7$, cutting hallucination $6.2\%\to1.6\%$ and missing steps
$3.3\%\to0.7\%$ and raising execution to $97.7$. On VH SD, where the
automaton already pins the answer down, the measure adds nothing, exactly as
this decomposition predicts. Representative masking outputs appear in
Appendix~\ref{apx:lookahead-examples}.

\textbf{The LM and the automaton are both load-bearing.} Search over the
compiled automaton with no LM is already strong ($83.8$ on VH AS), an honest
measure of how much structure the constraints carry; the LM contributes
grounding and ordering common sense worth $+6.9$ on VH AS and $+16$ on VH SD,
and the failures it removes are groundings no person would pick. In the other
direction, a classical planner given the benchmark's own full PDDL domain,
strictly more information than our automata encode, solves nearly every
problem it is posed yet scores $13$/$37$ on BEH AS and $47.2$/$84.6$ on VH AS:
wherever its model or goal language falls short of the evaluated environment
it returns confidently wrong plans, while encoding constraints rather than a
world model degrades gracefully when knowledge is incomplete.

\textbf{The trained surrogate is the full model the derivation calls for;
uniform is its zero-training approximation.} The trained HMM approximates the
LM's own distribution over futures, and it delivers the best execution SR on
three of the four modules. On task success the uniform surrogate matches or
slightly exceeds it, and the reason is a measurable
bias rather than an equivalence: this benchmark's mandatory but rare actions
almost never appear in the LM's unconstrained outputs (\textsc{Sleep} and
\textsc{PlugIn} occur zero times in ${\sim}10^{8}$ sampled tokens), so a
faithful surrogate assigns them near-zero mass and fines exactly the steps
the constraints exist to enforce; the uniform surrogate is immune by
construction. Training thus pays where the automaton leaves many admissible
futures and the corpus covers them, and costs little where it does not;
Appendix~\ref{apx:hmm} quantifies both directions.

\subsection{Real Robot Demonstrations}
\label{sec:exp-robot}
\begin{figure}
    \centering
    \includegraphics[width=0.99\linewidth]{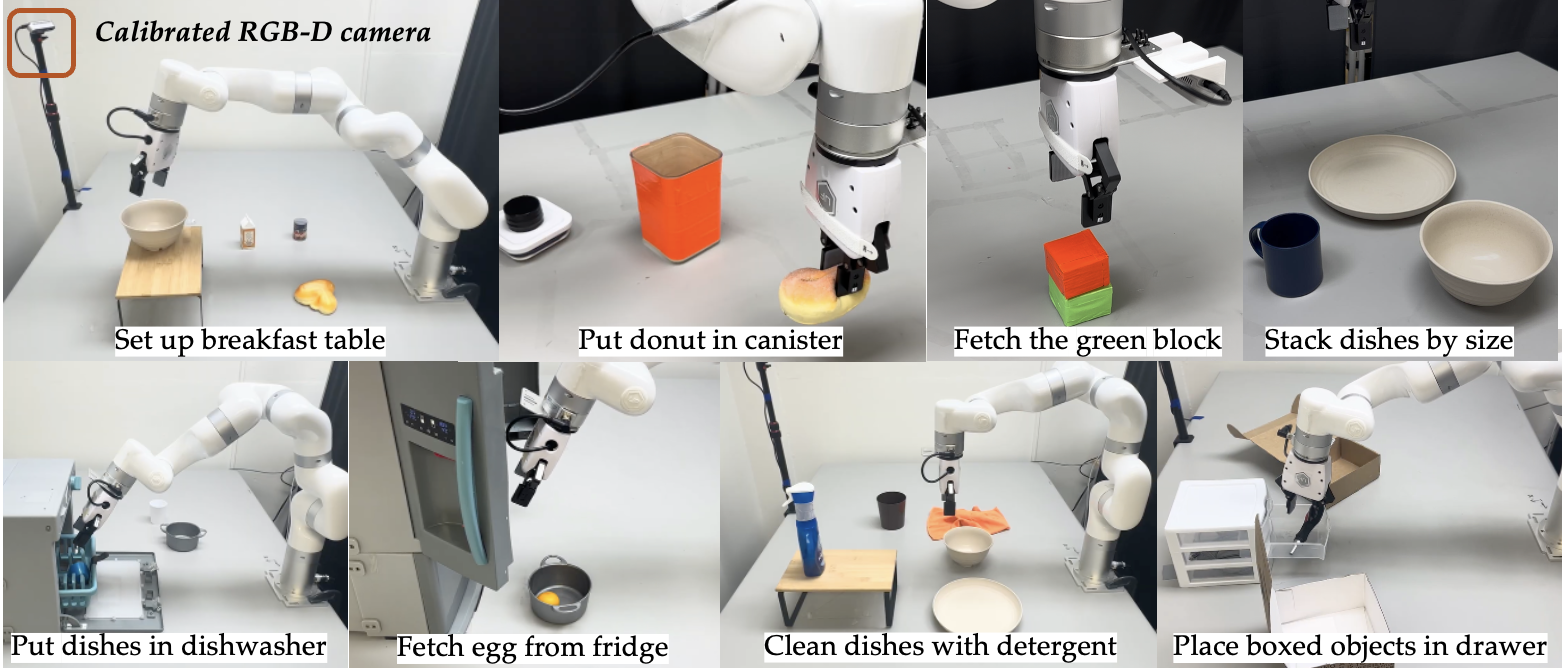}
    \caption{
    \small{Real robot setup with xArm 7 and deployed tabletop manipulation tasks. Top-row tasks require only pick-and-place; bottom-row tasks require specialized motion primitives (e.g. open door).}
    }
    \label{fig:real-tasks}
    \vspace{-1em}
\end{figure}
We deploy Meta-Ctrl on a tabletop XArm7 with a calibrated Realsense RGB-D camera, composing Meta-Ctrl plans with a Code-as-Policies executor~\cite{liang2023code} that grounds objects via Molmo~\cite{deitke2025molmo} and SAM2~\cite{ravi2025sam} and synthesizes grasps via GraspGen~\cite{murali2025graspgen}.
We demonstrate 8 BEHAVIOR-derived long-horizon manipulation tasks (Fig.~\ref{fig:real-tasks}) and quantitatively evaluate 3 pick-and-place tasks over 20 rollouts each.
\begin{table}[t]
\caption{%
  \small{
  Real robot evaluations on an XArm7, $20$ repetitions per task.
  Meta-Ctrl columns are a cascade: plan success holds by construction;
  perception and execution are conditional on preceding stages.
  Baseline perception is scored per-module, independent of plan
  validity; its execution is end-to-end and bounded by planning
  failures. Baseline downstream columns are thus not directly
  comparable to Meta-Ctrl's.}
  }
\label{tab:robot}
\centering
\small
\setlength{\tabcolsep}{4pt}
\begin{tabularx}{\linewidth}{l YYY YYY}
\toprule
& \multicolumn{3}{c}{\textbf{Meta-Ctrl (ours)}} & \multicolumn{3}{c}{\textbf{Baseline (Llama 3 8B)}} \\
\cmidrule(lr){2-4} \cmidrule(lr){5-7}
\textbf{Task} & \textbf{Planning} & \textbf{Perception} & \textbf{Execution} & \textbf{Planning} & \textbf{Perception} & \textbf{Execution} \\
\midrule
Put donut in canister  & 20/20 & 17/20 & 16/20 & 2/20 & 18/20 & 1/20 \\
Pick up green block (under blocker) & 20/20 & 19/20 & 18/20 & 1/20 & 19/20 & 1/20 \\
Stack dishes by size                 & 20/20 & 20/20 & 17/20 & 0/20 & 19/20 & 0/20 \\
\bottomrule
\end{tabularx}
\vspace{-1em}
\end{table}
Meta-Ctrl achieves $100\%$ plan validity by construction, including the multi-step precondition sequences needed for articulated containers. The unconstrained baseline routinely skips these preconditions---\textsc{Place} into a closed container, \textsc{Grasp} on an occluded object, or \textsc{Grasp} while holding another object---so its failures concentrate at planning. Meta-Ctrl's remaining failures occur downstream, at perception or grasp execution, isolating cleanly from planning errors. Representative rollouts and baseline precondition violations appear in Appendix~\ref{apx:real_demo}.

\section{Conclusion}
We presented \textbf{Meta-Ctrl}, a constrained-decoding framework
that guarantees LLM-generated plans satisfy their encoded syntactic and semantic constraints by decoupling enforcement across granularities via meta-tokens.
Across EAI, WAH-NL, and a real tabletop robot,
Meta-Ctrl turns a small open-weight LM into a competitive planner
that matches or exceeds models an order of magnitude larger,
with every output constraint-satisfying by construction.

\textbf{Limitations.} 
Meta-Ctrl assumes both constraints are DFA-expressible: syntax over
tokens, semantics over meta-tokens. Constraints requiring unbounded
memory (e.g., arbitrary arithmetic over continuous quantities) fall
outside this class. The semantic DFA is built per task family from a
structured specification; our experiments use the simulator's
predicates and, on BEHAVIOR, the benchmark's published object taxonomy
(Appendix~\ref{app:beh-affordance}). Domains without such a published
specification will need learned constraint extractors or more permissive
constraints, and the guarantee then holds relative to whatever the extractor
recovers. On a real system the automaton is initialised from the
perception-estimated symbolic state, so the guarantee is relative to the
robot's believed state rather than the true one; carrying a belief over
automaton states, rather than a single state, is the natural extension. The
guarantee is likewise relative to the encoded constraint, and an incomplete
specification can over-constrain: on BEHAVIOR our fixed affordance
prior rejects $17\%$ of gold plans (Appendix~\ref{app:beh-affordance}),
with infeasible tasks falling back to $\gamma$-only decoding.
Real-robot goal success is bounded by perception and grasping, not the
planner; closing that gap is orthogonal to plan generation. Finally,
HMM surrogates are shared across simulators (Appendix~\ref{apx:hmm})
but trained per module type; a single surrogate spanning module types
and base models remains future work.

\clearpage
\acknowledgments{This work was funded in part by the DARPA ANSR and CODORD programs under awards FA8750-23-2-0004 and HR00112590089, by DARPA SAFRON under grant HR0011-25-3-0141, and by gifts from Cisco Research, Qualcomm, and Amazon. Approved for public release; distribution is unlimited.}

\bibliography{example}

\appendix
\clearpage
\section{Construction of the Joint DFA $\dfa_\alpha$}
\label{apx:joint_dfa_construction}

We give the explicit construction of a token-level DFA $\dfa_\alpha$
that accepts the joint constraint
$\alpha(x) = \gamma(x) \wedge \beta(\tau(x))$
from Eq.~\ref{eq:alpha-def},
and bound its size by $|\dfa_\gamma| \cdot |\dfa_\beta|$.
This construction is the baseline that our factored approach
(Section~\ref{sec:method}) avoids.

\textbf{Setup.}
Let $\dfa_\gamma = (S_\gamma, V, \delta_\gamma, s_\gamma^0, F_\gamma)$
be the token-level syntax DFA over the LLM vocabulary $V$,
with states $S_\gamma$,
transition function $\delta_\gamma$,
initial state $s_\gamma^0$,
and accepting states $F_\gamma$.
Let $\dfa_\beta = (S_\beta, \mathcal{W}, \delta_\beta, s_\beta^0, F_\beta)$
be the meta-token-level semantic DFA over $\mathcal{W}$,
with the analogous components.
The parser $\tau : \mathrm{supp}_\gamma \to \mathcal{W}^*$
maps each completed action block in $\mathrm{supp}_\gamma$
to a single meta-token in $\mathcal{W}$.
Let $\mathcal{C}_\gamma \subseteq S_\gamma$
denote the subset of $\dfa_\gamma$ states
at which an action block has just been completed
(i.e., states reached by the token that closes the block).

\textbf{Construction.}
We construct $\dfa_\alpha$ as follows:
\begin{itemize}
\item \emph{States:} $S_\alpha = S_\gamma \times S_\beta$.
\item \emph{Initial state:} $(s_\gamma^0, s_\beta^0)$.
\item \emph{Transition function.}
  For each $(q_\gamma, q_\beta) \in S_\alpha$ and each token $v \in V$,
  let $q_\gamma' = \delta_\gamma(q_\gamma, v)$.
  If $q_\gamma'$ is undefined (the token violates syntax),
  the transition is undefined.
  Otherwise,
  \begin{itemize}
    \item If $q_\gamma' \in \mathcal{C}_\gamma$
      (token $v$ closed an action block),
      let $a \in \mathcal{W}$ be the meta-token produced by $\tau$
      from the block tokens; advance the semantic state:
      \[
        \delta_\alpha\bigl((q_\gamma, q_\beta), v\bigr)
          = \bigl(q_\gamma', \delta_\beta(q_\beta, a)\bigr).
      \]
    \item If $q_\gamma' \notin \mathcal{C}_\gamma$
      (token $v$ did not close a block),
      hold the semantic state fixed:
      \[
        \delta_\alpha\bigl((q_\gamma, q_\beta), v\bigr)
          = \bigl(q_\gamma', q_\beta\bigr).
      \]
  \end{itemize}
\item \emph{Accepting states:}
  $F_\alpha = F_\gamma \times F_\beta$.
\end{itemize}

\textbf{Correctness.}
By induction on $|x|$, for any $x \in V^*$,
$\dfa_\alpha$ reaches state $(q_\gamma, q_\beta)$
if and only if
$\dfa_\gamma$ reaches $q_\gamma$ on $x$ and
$\dfa_\beta$ reaches $q_\beta$ on $\tau_{\mathrm{inc}}(x)$.
Hence $x \in L(\dfa_\alpha)$ iff $x \models \gamma$
(reaches $F_\gamma$) and $\tau(x) \models \beta$
(reaches $F_\beta$),
matching Eq.~\ref{eq:alpha-def}.

\textbf{Size bound.}
The state space satisfies
$|S_\alpha| \le |S_\gamma| \cdot |S_\beta|$,
and each state has at most $|V|$ outgoing transitions,
giving
\[
  |\dfa_\alpha| \;\le\; |S_\gamma| \cdot |S_\beta| \cdot |V|
  \;\le\; |\dfa_\gamma| \cdot |\dfa_\beta|
\]
when DFA size is measured in edges and $|V|$ is absorbed into
the per-state branching.
Reachability and minimization can reduce this constant,
but the multiplicative scaling persists in the worst case
because each state in $\mathcal{C}_\gamma$ must branch
into $|S_\beta|$ distinct successors,
one per reachable semantic configuration.
This is the bottleneck our factored approach avoids:
we never construct $\dfa_\alpha$,
only $\dfa_\gamma$ and $\dfa_\beta$ separately.

\newpage

\section{Two-Level Decoding Algorithm}
\label{apx:algorithm}
\begin{algorithm}[h]
\caption{Two-Level Constrained Decoding}
\label{alg:two-level}
\KwIn{Prompt $x_{\mathrm{prompt}}$;
  LM $p_{\mathrm{LM}}$;
  syntax HMM and DFA $\dfa_\gamma$;
  meta-token HMM and DFA $\dfa_\beta$;
  parser $\tau$ (incremental form $\tau_{\mathrm{inc}}$).}
\KwOut{Token sequence satisfying $\alpha = \gamma \wedge (\beta \circ \tau)$.}
\BlankLine
$B_\gamma \gets$ backward DP over $\mathrm{HMM}_{\mathrm{token}} \times \dfa_\gamma$
  \tcp*{token-level lookahead}
$B_\beta \gets$ backward DP over $\mathrm{HMM}_{\mathrm{meta}} \times \dfa_\beta$
  \tcp*{action-level lookahead}
\BlankLine
$x \gets x_{\mathrm{prompt}}$;\quad $l \gets 1$\;
\Repeat{$x_t = \texttt{<EOS>}$}{
  $a_{<l} \gets \tau_{\mathrm{inc}}(x)$
    \tcp*{completed actions so far}
\For{each token $v \in V$ admissible at the current $D_\gamma$ state}{
    $q(v) \leftarrow p_{\text{LM}}(v \mid x) \cdot p(\alpha \mid x{\cdot}v)$
    \tcp*{constraint-conditioned posterior, Eq.~\ref{eq:final}}
}
$x_t \leftarrow \arg\max_v q(v)$;\quad $x \leftarrow x \cdot x_t$;
  \If{$x_t$ closes an action block}{$l \gets l+1$}
}
\Return $x$
\end{algorithm}

\section{Per-Module DFA and Parser Construction}
\label{apx:dfas}

We build a separate syntactic DFA $\dfa_\gamma$, semantic DFA $\dfa_\beta$,
and meta-token parser $\tau$ for each (benchmark, module) pair.
All four core modules share the same three-part skeleton:
$\dfa_\gamma$ is a deterministic state machine over the base-LM token
alphabet that admits exactly the grounded action$/$predicate plus
type-compatible object-instance pairs derivable from the prompt;
$\dfa_\beta$ is a state machine over the meta-token alphabet that rejects
on a precondition violation and accepts when every (non-dropped) task
goal is satisfied; $\tau$ consumes the same token stream as $\dfa_\gamma$
and emits one meta-step per completed action$/$atom boundary, remapping
the prompt's raw object ids to contiguous $0$-based canonical ids.

\subsection{VirtualHome Action Sequencing (VH-AS)}
\label{apx:dfas:vh-as}

\paragraph{Output grammar ($\dfa_\gamma$).}
Plans are emitted as a single JSON object whose keys are action names and
whose values are flat argument arrays \texttt{[name, id]} (or
\texttt{[name$_0$, id$_0$, name$_1$, id$_1$]} for two-argument actions):
\begin{quote}\small\ttfamily
\{\,"WALK": ["soap", "1002"]\,,\\
\hphantom{\{}"GRAB": ["soap", "1002"]\,,\\
\hphantom{\{}"PUTON": ["clothes\_jacket", "1003"]\,\}
\end{quote}
$\dfa_\gamma$ threads three per-scene tries (action-name, object-name,
and per-name id) compiled once from the prompt's object listing.
After a complete action block, the DFA returns to its block-start
state; an arbitrary-length action list is emitted until the close brace
followed by the end-of-turn token. Actions are offered only when all
of their argument roles have at least one type-compatible object in the
scene.

\paragraph{Action vocabulary ($\mvocab$).}
The VH-AS action set has $|\mvocab_{\text{VH-AS}}| = 36$ types:
one nullary (\textsc{StandUp}); three binary (\textsc{Pour},
\textsc{PutBack}, \textsc{PutIn}); and $32$ unary actions including
\textsc{Walk}, \textsc{Find}, \textsc{Grab}, \textsc{Open},
\textsc{SwitchOn}, \textsc{PlugIn}, \textsc{Sit}, \textsc{Drink},
\textsc{Cut}, \textsc{Read}, \textsc{Type}, \textsc{Wash},
\textsc{Scrub}, \textsc{Wipe}.
The set is a subset of the official evaluator's valid actions, omitting
those that are non-executable or never goal-bearing.

\paragraph{Semantic constraints ($\dfa_\beta$).}
$\dfa_\beta$ tracks a state tuple consisting of: agent location,
both hand-occupancy slots, posture, facing direction, and per-object
bitmasks for openable, switchable, and plugged-in state, plus
relational bits for graph edges and ordered action goals.
Static preconditions are derived from prompt-stated object properties:
\textsc{Grab} requires \texttt{GRABBABLE}; \textsc{Open}$/$\textsc{Close}
require \texttt{CAN\_OPEN}; \textsc{SwitchOn}$/$\textsc{SwitchOff}
require \texttt{HAS\_SWITCH}; \textsc{PlugIn}$/$\textsc{PlugOut} require
\texttt{HAS\_PLUG}; \textsc{Sit}$/$\textsc{Lie} require
\texttt{SITTABLE}$/$\texttt{LIEABLE}; container$/$surface arguments of
\textsc{PutIn}$/$\textsc{PutBack} require \texttt{CONTAINERS}$/$
\texttt{SURFACES}. Dynamic preconditions are checked against the live
state: a free hand and proximity for \textsc{Grab}; plugged-in and
not-open for \textsc{SwitchOn}; not-on for \textsc{Open}; held object
plus open destination for \textsc{Put*}; posture for locomotion.
Goals are partitioned into node, edge, character-state, and ordered
action goals; a state is accepting iff all (non-dropped) goals hold.

\paragraph{Parser ($\tau$).}
$\tau$ advances in lock-step with $\dfa_\gamma$ and emits one meta-step
at each block-boundary closure (the \texttt{]} closing an action's
argument array), reading the action name and raw ids from trie leaves
and remapping each raw id to canonical form. 

\subsection{VirtualHome Subgoal Decomposition (VH-SD)}
\label{apx:dfas:vh-sd}

\paragraph{Output grammar ($\dfa_\gamma$).}
Plans are a JSON object with a single \texttt{"output"} key whose value
is a list of predicate atoms in temporal order:
\begin{quote}\small\ttfamily
\{"output": ["PLUGGED\_IN(washing\_machine.1001)",\\
\hphantom{\{"output": [}"CLOSED(washing\_machine.1001)",\\
\hphantom{\{"output": [}"ON(washing\_machine.1001)"]\}
\end{quote}
Each atom is \texttt{PRED(name.id)} or \texttt{PRED(name$_0$.id$_0$,
name$_1$.id$_1$)} with arguments written in dotted form. Cross-argument
predicates carry forward the chosen \texttt{arg$_0$} so that
\texttt{arg$_1$}'s admissible set is filtered by \texttt{arg$_0$}'s
identity.

\paragraph{Subgoal vocabulary ($\mvocab$).}
$|\mvocab_{\text{VH-SD}}| = 36$ predicates: $9$ state
(\textsc{Closed}, \textsc{Open}, \textsc{On}, \textsc{Off},
\textsc{Plugged\_In}, \textsc{Plugged\_Out}, \textsc{Dirty},
\textsc{Sitting}, \textsc{Lying}); $6$ relation (\textsc{Inside},
\textsc{OnTop}, \textsc{Next\_To}, \textsc{Facing}, \textsc{Holds\_LH},
\textsc{Holds\_RH}); and $21$ action-named intermediate-subgoal
predicates (\textsc{Grab}, \textsc{Clean}, \textsc{Cut}, etc.).

\paragraph{Semantic constraints ($\dfa_\beta$).}
$\dfa_\beta$ carries per-object state bitmasks plus goal-progress bits.
Executor ordering is enforced as predicate preconditions:
\textsc{On} requires \textsc{Plugged\_In} and not-\textsc{Open};
\textsc{Open} requires not-\textsc{On};
\textsc{Inside}$/$\textsc{OnTop} require the destination container
\textsc{Open};
\textsc{Holds}$_{LH/RH}$ require a free hand and a grabbable target.
State and placement goals are evaluated from the live state at accept
time (re-emitting \textsc{Open} after \textsc{Closed} correctly leaves
the object open), and repeated placements use latest-wins semantics
mirroring the evaluator. A state is accepting iff every (non-dropped)
goal is currently satisfied.

\paragraph{Parser ($\tau$).}
$\tau$ emits one meta-step per completed atom, reading the predicate
name and dotted \texttt{name.id} arguments from trie leaves and
remapping ids. Structural tokens emit nothing.

\subsection{BEHAVIOR Action Sequencing (BEH-AS)}
\label{apx:dfas:beh-as}

\paragraph{Output grammar ($\dfa_\gamma$).}
Plans are a JSON array of one-action dictionaries
\texttt{\{"action": NAME, "object": INSTANCE\}}:
\begin{quote}\small\ttfamily
[\{"action": "LEFT\_GRASP", "object": "plywood\_0"\},\\
\hphantom{[}\{"action": "SLICE", "object": "strawberry\_0"\},\\
\hphantom{[}\{"action": "LEFT\_PLACE\_INSIDE", "object": "jar\_0"\}]
\end{quote}
BEHAVIOR object instances are written as \texttt{base\_name\_index}.
$\dfa_\gamma$ recognizes the action via a vocabulary trie and the
instance via a per-scene instance trie built from the prompt, closing
the list with \texttt{]} and EOS. As in VH-AS, the grammar is cyclic
across action blocks and applies viable-action and affordance pruning
(Appendix~\ref{app:beh-affordance}).

\paragraph{Action vocabulary ($\mvocab$).}
$|\mvocab_{\text{BEH-AS}}| = 30$ action types, all hand-explicit where
the simulator is bimanual: left$/$right \textsc{Grasp},
\textsc{Release}, \textsc{Place\_OnTop}, \textsc{Place\_Inside},
\textsc{Place\_NextTo}, \textsc{Place\_Under},
\textsc{Place\_NextTo\_OnTop} (binary), and
\textsc{Transfer\_Contents\_Inside}$/$\textsc{OnTop}; plus the
hand-agnostic state actions \textsc{Open}, \textsc{Close},
\textsc{Toggle\_On}, \textsc{Toggle\_Off}, and the attribute actions
\textsc{Clean}, \textsc{Slice}, \textsc{Soak}, \textsc{Dry},
\textsc{Cook}, \textsc{Freeze}, \textsc{Unfreeze}. Navigation is
implicit and is not emitted.

\paragraph{Semantic constraints ($\dfa_\beta$).}
$\dfa_\beta$ carries goal bits, two hand-occupancy slots, a per-object
location hash, and per-object boolean state bits. It encodes both
task-specific preconditions parsed from the prompt's initial-state
block and the task-independent affordance prior of
Appendix~\ref{app:beh-affordance}: \textsc{Grasp} requires a free
target hand, a graspable target, and no closed container in the
target's containment ancestry; \textsc{Place\_Inside} requires the
destination to be open; \textsc{Toggle\_On} is forbidden on an open
appliance; \textsc{Soak} requires the target inside a toggled-on water
source; \textsc{Clean} of a stained target requires holding a soaked
cleaner. A state accepts iff all (non-dropped) goal bits are set.

\paragraph{Parser ($\tau$).}
$\tau$ emits one meta-step when each per-action dictionary closes,
mapping captured \texttt{base\_name\_index} strings to canonical ids.

\subsection{BEHAVIOR Subgoal Decomposition (BEH-SD)}
\label{apx:dfas:beh-sd}

\paragraph{Output grammar ($\dfa_\gamma$).}
Plans are a JSON object \texttt{\{"output": [\,\dots\,]\}} whose
elements are temporally ordered subgoal steps; within a step, atoms
are joined by \texttt{" and "} (a conjunctive set):
\begin{quote}\small\ttfamily
\{"output": ["holds\_rh(carton.66) and inside(hardback.64, carton.66)\\
\hphantom{\{"output": [}\ and inside(hardback.65, carton.66)", \dots]\}
\end{quote}
Each atom is \texttt{[not ]pred(name.id[, name.id])} with lowercase
predicate names. The grammar uses \texttt{" and "} within a step and
\texttt{", "} between steps.

\paragraph{Subgoal vocabulary ($\mvocab$).}
$16$ positive predicates plus their \texttt{NOT\_} negations ($32$
atoms total): $8$ state (\texttt{open}, \texttt{toggledon},
\texttt{cooked}, \texttt{frozen}, \texttt{sliced}, \texttt{soaked},
\texttt{stained}, \texttt{dusty}); $8$ relation (\texttt{inside},
\texttt{ontop}, \texttt{under}, \texttt{onfloor}, \texttt{nextto},
\texttt{touching}, \texttt{holds\_lh}, \texttt{holds\_rh}).
Anti-goals are first-class.

\paragraph{Semantic constraints ($\dfa_\beta$).}
$\dfa_\beta$ carries goal bits, container-open bits, and a clean$/$soak
chain bitmask. It accepts every $\gamma$-valid atom, setting the
matching goal bit on positive match. Two precondition mechanisms are
retained. \emph{Gate}: an object the prompt's initial state places
inside a closeable container requires that container's open bit before
any non-\texttt{open} predicate may target it.
\emph{Clean$/$soak chain}: \texttt{soaked} or \texttt{not stained}
goals are creditable only after the place--toggle--soak prerequisite
chain. Goals are obtained by expanding the prompt's quantified goal
AST; anchored relations whose anchor is itself repositioned impose
the ordering that the anchor's placement goal must be set first.

\paragraph{Parser ($\tau$).}
$\tau$ emits one meta-step per atom, mapping instance labels to
canonical ids.

\subsection{WAH-NL via LoTa-Bench}
\label{apx:dfas:wah}

The WAH-NL setting reuses the VirtualHome machinery with
Llama-3.1-8B-Instruct as the base LM. $\dfa_\gamma$ is the VH-AS DFA
restricted to the WAH-relevant action subset (navigation,
manipulation, object interaction, posture; \textsc{WakeUp},
\textsc{Sleep}, and \textsc{PutObjBack} are trimmed). $\dfa_\beta$ is
the VirtualHome state machine of Appendix~\ref{apx:dfas:vh-as}
applied to WAH goals: \textsc{Grab} requires a free hand and
co-location; \textsc{PutIn} requires holding the object and the
destination container not closed; \textsc{Open}$/$\textsc{Close}
require co-location; \textsc{SwitchOn} requires the appliance
plugged-in. These are the invariants whose violation accounts for the
$60$\,/\,$21$\,/\,$10$ failure distribution of the ProgPrompt
baseline in Section~\ref{sec:exp-wah}: any action whose precondition
fails is never offered to the LM. $\tau$ is the VirtualHome parser of
Appendix~\ref{apx:dfas:vh-as} applied to the WAH action subset.

\subsection{Shared construction notes}
\label{apx:dfas:shared}

\paragraph{Canonical tokenization.}
All DFAs operate over the base LM's token alphabet using a single
canonical tokenization per atom. Token sets are computed once per
tokenizer; structural substrings with more than one tokenization are
treated as equivalence classes and marginalized over by
\texttt{logsumexp}, so no tokenization variant collapses the candidate
set mid-stream.

\paragraph{EOS handling.}
End-of-sequence is treated as a meta-token gated by $\beta$: EOS is
admitted only at an action$/$atom boundary, and only when the $\beta$
state is accept-reachable.

\section{HMM Training}
\label{apx:hmm}

\paragraph{Training data construction.}
For each (benchmark, module) pair we construct a training set of
unconstrained base-LM generations whose prompts share the benchmark
format but contain no evaluation task instances. Specifically, we hold
out the official evaluation tasks and construct training prompts by
category-preserving substitution: object names and predicate arguments
in the EAI/WAH-NL prompt template are resampled from the corresponding
vocabularies (objects, properties, locations) so the surface format and
predicate distribution match evaluation but the actual instances do not.
We sample up to $1{,}000{,}000$ generations per (benchmark, module) pair at
temperature $0.1$, top-$p$ $0.95$, length cap $2048$ tokens (seed $42$); the
BEHAVIOR halves comprise $492{,}352$ (AS) and $703{,}360$ (SD) raw
generations. Generations that fail to parse are discarded
($\gamma$-acceptance for AS outputs, closed-output for SD), and the two
simulators' surviving halves are subsampled to a balanced $1{:}1$ mixture
before training. The retained sequences are used for both the token-level HMM
(raw token sequence) and the meta-token HMM (the $\tau$-projection of the same
sequence).

\paragraph{Corpus diversity.}
Continuations are sampled with a distinct seed per generation call. At
temperature $0.1$ the corpora are highly redundant: BEH AS yields $58$
distinct meta-token sequences from $95{,}665$ samples over $70$ prompts, and
scaling VH AS by $5.3\times$ adds $14$; held-out NLL agrees ($9$--$10\times$
more data buys $0.07$--$0.08$ nats, and the last $1.6\times$ buys $0.005$).
The temperature-$1.0$ corpus is far more diverse ($294$ of $512$ raw
continuations distinct per prompt; $34.8\%$ of meta-token sequences distinct
on BEH AS). Diversity does not, however, recover mandatory-but-rare actions,
which are absent from the base LM's output distribution at any temperature.

\textbf{HMMs.} For EAI we train two meta-token HMMs: a 132-symbol Action Sequencing HMM shared by
VH-AS and BEH-AS, trained on the union of the two simulators' AS augmented generations, and a
123-symbol Subgoal Decomposition HMM shared by VH-SD and BEH-SD, trained on the union of their SD
generations. WAH-NL uses a third meta-token HMM. Each uses hidden dimension $H{=}128$. We train one
token-level HMM per base LM over that model's tokenizer vocabulary (Llama 3: 128K, gpt-oss: 200K)
and reuse it unchanged across all of that base LM's modules and simulators. The meta-token alphabet
decomposes into grounded-action (or predicate) names, one object-ID symbol per argument slot,
\textsc{Unk}, and \textsc{Eos}:\footnote{A meta-token is one grounded action, the symbol over which
$D_\beta$ is defined (e.g.\ \textsc{Grab}(plate)). The HMM emits each grounded action as its name
followed by one object-ID symbol per argument: a nullary action is one symbol, a unary action two
([\textsc{action}, ID]), a binary action three ([\textsc{action}, ID, ID]). Hand assignment is
folded into the action name (e.g.\ \textsc{Left\_Grasp}), so it carries no separate symbol. The
vocabulary sizes count these emission symbols, not distinct grounded actions.} for Action
Sequencing the 132 symbols are 81 action names, 49 object-ID symbols (\texttt{ID0} through
\texttt{ID48}), \textsc{Unk}, and \textsc{Eos}; Subgoal Decomposition's 123 decompose analogously.
The main configuration uses the trained meta-HMM; the uniform variant below is an ablation.

\begin{table}[h]
\caption{HMM parameter counts. Token-level HMM size is dominated by the
emission matrix $H \times |V|$; meta-token HMM is dominated by the
$H \times H$ transition matrix at this scale. $H{=}128$ throughout; token-level
$|V|$ shown for the Llama-3 tokenizer (gpt-oss uses $|V|{\approx}201$K $\Rightarrow$ ${\sim}25.7$M).}
\label{tab:hmm-params}
\centering
\small
\begin{tabular}{lrrr}
\toprule
& \textbf{Vocab} & \textbf{Token-level HMM} & \textbf{Meta-token HMM} \\
\midrule
VH-AS  & 132 actions / $128$K tokens & ${\sim}16.4$M & ${\sim}33$K \\
VH-SD  & 123 predicates / $128$K     & ${\sim}16.4$M & ${\sim}32$K \\
BEH-AS & 132 actions / $128$K        & ${\sim}16.4$M & ${\sim}33$K \\
BEH-SD & 123 predicates / $128$K     & ${\sim}16.4$M & ${\sim}32$K \\
WAH-NL & 7 actions + 128 ids / $128$K & ${\sim}16.4$M & ${\sim}34$K  \\
\bottomrule
\end{tabular}
\end{table}

\paragraph{Trained vs.\ uniform meta-HMM.}
We compare the trained meta-token HMM against a uniform variant in which
the emission and transition parameters are uniform (i.e., no learned
action-level prior; the meta-HMM contributes no preference beyond which
actions are constraint-reachable and how far they are from acceptance). The
main-table configuration uses the trained meta-HMM. Both surrogates enforce
the same constraint and carry the same guarantee; they differ only in how
they rank the constraint-satisfying set.

\begin{center}
\small
\captionof{table}{Trained vs.\ uniform meta-token HMM on Llama~3 8B, task /
execution SR (\%). All trained rows use the single recipe of
\cref{apx:hmm}: parse-filtered corpus, balanced 1:1 simulator mixture, App.~D
hyperparameters, no decode-time scaling. The last row retrains the same
recipe on a temperature-$1.0$ corpus. VH-AS cells are means over six
evaluator seeds (Appendix~\ref{apx:evalnoise}).}
\label{tab:trained-vs-uniform}
\begin{tabular}{lcccc}
\toprule
\textbf{Meta-HMM} & \textbf{VH AS} & \textbf{VH SD} & \textbf{BEH AS} & \textbf{BEH SD} \\
\midrule
Uniform                    & \textbf{90.7}~/~97.7 & 88.4~/~93.2 & \textbf{36}~/~89 & \textbf{41}~/~59 \\
Trained ($T{=}0.1$ corpus) & 90.4~/~97.7 & \textbf{89.9}~/~\textbf{93.5} & 34~/~89 & 36~/~\textbf{66} \\
Trained ($T{=}1.0$ corpus) & 90.1~/~\textbf{98.0} & 87.8~/~92.6 & 33~/~\textbf{96} & \textbf{42}~/~65 \\
\bottomrule
\end{tabular}
\end{center}

\noindent
The two configurations trade places by a few points per module. The trained
surrogate wins where the automaton leaves many admissible continuations and
the corpus covers them: with the more diverse temperature-$1.0$ corpus it
overtakes uniform on BEH SD ($42$ vs $41$) and takes the best execution SR on
the action-sequencing modules; the uniform surrogate is slightly ahead
elsewhere. Where the trained surrogate loses, the mechanism is a bias toward
corpus typicality: this benchmark's mandatory but rare actions almost never
appear in the LM's unconstrained outputs (\textsc{Sleep} and \textsc{PlugIn}
never occur in ${\sim}10^{8}$ sampled tokens), so a faithful surrogate assigns
them near-zero mass and fines exactly the steps the constraints exist to
enforce; the uniform surrogate is immune by construction. A fuller treatment
of surrogate calibration for constrained decoding is left to future work.


\section{Decoding Details}
\label{apx:decoding}

\paragraph{Greedy decoding under the factored posterior.}
At each decoding step we evaluate the factored posterior of
Eq.~\ref{eq:final} and select the argmax token. We do not use
temperature sampling, top-$p$, or beam search; the constraint-aware
reweighting is the only departure from greedy.

\paragraph{Decoding overhead.}
At batch size $1$, constrained decoding runs at parity with unconstrained
greedy decoding ($20.6$ vs $20.9$ ms per token, medians over $12$ VH-AS
tasks with matched generation lengths); per-task automaton construction adds
a median $2.3$\,s ($\gamma$) and $0.7$\,s ($\beta$), and peak device memory
is $32$\,GiB including the fp16 8B model.

\paragraph{Horizon budget.}
The semantic backward DP requires a finite horizon over meta-tokens.
We set this budget to $40$ actions for all EAI and WAH-NL evaluations.
This bound is generous: across the four EAI modules, gold plans have a
median length of $4$ meta-tokens and a $95$th-percentile length of $19$,
well within budget. Only a single VH AS task ($54$-step plan) exceeds
the cap; remaining 879 remaining evaluation tasks fit. The budget cap is
not the primary source of fallback.

\section{Qualitative Examples: DFA-only Degeneration vs.\ Meta-Ctrl}
\label{apx:lookahead-examples}

Section~\ref{sec:exp-eai} reports that hard masking on the same
$\gamma$ DFA collapses task success while inflating execution rate.
DFA-only collapses to a single trivially-true predicate after the LM's
first sample, because no token in the local mask is preferred over
closing the bracket.
Meta-Ctrl emits a concise, goal-completing plan because the
$\beta$ backward DP assigns zero mass to any continuation that closes
the bracket before the goal predicates are reached.

\begin{table}[h]
\caption{%
  Representative outputs from DFA-only ($\gamma$ hard mask) and
  Meta-Ctrl ($\gamma{+}\beta$) on VirtualHome Subgoal Decomposition.
  DFA-only emits a single locally-true predicate and stops;
  Meta-Ctrl produces the predicates required to reach the goal.}
\label{tab:degeneration-examples}
\centering
\small
\begin{tabular}{p{0.16\textwidth} p{0.14\textwidth} p{0.28\textwidth} p{0.32\textwidth}}
\toprule
\textbf{Task} & \textbf{Prompt id} & \textbf{DFA-only ($\gamma$ mask)} & \textbf{Meta-Ctrl ($\gamma{+}\beta$)} \\
\midrule
Turn on light & scene\_1\_125\_2 &
  \texttt{NEXT\_TO(character.65, light.411)} &
  \texttt{PLUGGED\_IN(light.411), ON(light.411)} \\
\bottomrule
\end{tabular}
\end{table}

\noindent
The DFA-only output is locally valid: \texttt{NEXT\_TO(character, light)}
is a syntactically correct subgoal predicate and is true in the initial
scene, so the executor accepts it and reports the plan as executed
($96.0\%$ execution rate on VH SD). It does not change the light's
state, so it does not satisfy the task goal ($0.0\%$ task SR on VH SD).
Meta-Ctrl's lookahead, scoring each token by the probability that the
remaining sequence satisfies $\beta$, rules out the early bracket close
and forces the LM to emit the predicates the goal requires.

\section{BEHAVIOR Affordance Prior}
\label{app:beh-affordance}

The BEHAVIOR system prompt specifies the action grammar, object
instances, initial-state relations, and goal predicates, but does not
expose per-object affordances (which objects are openable, toggleable,
graspable, valid containers, or valid cleaning agents). VirtualHome
prompts carry this information directly via property tags; the
BEHAVIOR prompt format does not. To match VirtualHome's setup, we
recover the missing affordance information from the benchmark's own
public object taxonomy.

\paragraph{Affordances come from the BDDL taxonomy, not from an author list.}
Each BEHAVIOR prompt names its object instances by WordNet synset
(e.g.\ \texttt{bucket.n.01}), and the BEHAVIOR Domain Definition Language
(BDDL)~\cite{srivastava2022behavior} ships a taxonomy over those synsets:
an ISA tree of $1{,}822$ nodes together with per-synset ability annotations
(\texttt{openable}, \texttt{toggleable}, \texttt{soakable}, \texttt{cleaningTool},
and so on). BDDL is the language in which the benchmark's own activity
definitions are written and is distributed in the public \texttt{bddl}
package; our copy is semantically identical to the released one
($1{,}822$ nodes, zero ability differences). Affordance membership for an
instance is therefore a lookup: the prompt supplies the synset, the taxonomy
supplies the abilities and the ISA class, and $D_\beta$ reads them off. No
substring matching over instance names and no author-curated keyword list is
involved.

\paragraph{Two additions on top of the taxonomy.}
The taxonomy is not quite sufficient, and we state exactly where we go beyond
it. First, an admissibility rule used at DFA-construction time: \textsc{Open}
and \textsc{Close} (and the corresponding \texttt{open} atoms in Subgoal
Decomposition) enter the action universe for an object only if the prompt's
initial or goal state mentions that object's open state, or the object is an
open-gated container in container \emph{role} for this task, i.e.\ the target
of an \texttt{inside} goal or the holder of an initial \texttt{inside}
relation. This is a statement about which actions the task gives evidence for,
not a generation-time heuristic, and it is load-bearing: removing it flips
seven tasks from success to failure and none in the other direction, worth
$+3$ task points on both BEHAVIOR modules. Second, five generic rules, each
justified by a statement about the physical world that stands without
reference to the evaluator: open-top vessels (\texttt{bucket}, \texttt{basket},
\texttt{bin} ISA subtrees) have no lid and are not open-gated; bathtubs are
tap-equipped water fixtures, hence toggleable water sources; kettles, teapots
and pools dispense or hold water without a switch, hence passive water
sources; an object a goal asks to be repositioned is graspable, whatever its
ISA class; and soaking or drying applies to cleaning implements unless the
prompt states otherwise. The first of these knowingly disagrees with the
benchmark's gold plans, which do emit \texttt{open} on buckets and baskets;
we keep the physical statement and absorb the resulting over-constraint.

\paragraph{General precondition rules.}
The $\beta$ state machine enforces a small number of task-independent
precondition rules, each a domain-general property of the simulator:
a grasp is rejected if the target hand is already holding an object;
interacting with or placing into a closeable container requires it to
be open; an object cannot be placed inside itself or anything
(transitively) contained in it; and a position anchor must be
established before relations referencing it.

\paragraph{Scope of the prior.}
The prior encodes the kinds of object-affordance facts that
VirtualHome's prompt format exposes directly. The taxonomy lookup, the one
admissibility rule, the five generic rules and the ${\approx}4$ precondition
families apply to every task in the module; none of them depends on a task's
goal, gold plan, or intermediate state. The information available to
$D_\beta$ is thus the prompt plus a published object taxonomy, strictly
less than the planner baselines of \cref{tab:lookahead-ablation} receive,
since the benchmark itself distributes a full BEHAVIOR PDDL domain with $30$
action schemas carrying preconditions \emph{and} effects, whereas $D_\beta$
encodes no effects.

\textbf{Over-constraint on gold plans.} Because the prior is fixed and task-independent, it can be
stricter than a given task requires. Applied to the BEHAVIOR gold plans with no LM in the loop, it
renders 17 of 100 infeasible, each containing one action the gold plan performs but a precondition
rule rejects. Meta-Ctrl falls back to $\gamma$-only decoding on these, so they cap achievable
BEHAVIOR task SR independently of the base LM. That the prior rejects this many gold plans is
itself evidence it is generic rather than fit to the gold set.

\section{Format and Hallucination Errors: Per-Module Breakdown}
\label{apx:format-errors}

Table~\ref{tab:format-errors} reports the rate of parsing, hallucination,
and parameter errors for the raw Llama~3 8B baseline and for Meta-Ctrl
across all four EAI (simulator, module) pairs. \emph{Total} is the
fraction of plans containing the corresponding error type. We then
decompose this total into a \emph{benchmark-side} component (errors
caused by quirks of the evaluator's input handling, where the LM's
output is in fact admissible under the prompt's stated grammar and
vocabulary) and an \emph{LM-fault} component (residual errors that would
constitute genuine constraint violations Meta-Ctrl was responsible for
preventing). The LM-fault column is uniformly zero by construction:
every Meta-Ctrl output satisfies its syntactic DFA $\dfa_\gamma$, which
encodes exactly the parsing and vocabulary constraints these error
categories test.

\begin{table}[h]
\caption{%
  Format and hallucination error rates for raw Llama~3 8B and
  Meta-Ctrl. \emph{Parsing}: output fails the evaluator's parser.
  \emph{Hallucination}: output references objects or predicates not
  present in the evaluator's per-task vocabulary. \emph{Parameter}:
  argument arity, type, or formatting mismatch. The Meta-Ctrl
  \emph{benchmark-side} column counts plans rejected by the evaluator
  even though they comply with the prompt's stated grammar and
  vocabulary (footnotes identify each case). The \emph{LM-fault} column
  is the residual after removing benchmark-side rejections, and is zero
  across every module and error class.}
\label{tab:format-errors}
\centering
\small
\begin{tabular}{llcccc}
\toprule
\textbf{Module} & \textbf{Error class} & \textbf{Raw Llama 3 8B} & \multicolumn{3}{c}{\textbf{Meta-Ctrl}} \\
\cmidrule(lr){4-6}
 & & & \textbf{Total} & \textbf{Benchmark-side} & \textbf{LM-fault} \\
\midrule
\multirow{4}{*}{BEH AS}
  & Parsing                 & 0.0\%  & 0.0\%  & 0.0\% & 0.0\% \\
  & Hallucination           & 15.0\% & 0.0\%  & 0.0\% & 0.0\% \\
  & Parameter               & 9.0\%  & 0.0\%  & 0.0\% & 0.0\% \\
  & \textbf{Sum format/hallucination} & \textbf{24.0\%} & \textbf{0.0\%} & \textbf{0.0\%} & \textbf{0.0\%} \\
\midrule
\multirow{4}{*}{VH AS}
  & Parsing                 & 14.4\% & 1.3\%  & 1.3\%\textsuperscript{a} & 0.0\% \\
  & Hallucination           & 6.9\%  & 1.3\%  & 1.3\%\textsuperscript{b} & 0.0\% \\
  & Parameter               & 1.0\%  & 0.0\%  & 0.0\% & 0.0\% \\
  & \textbf{Sum format/hallucination} & \textbf{22.3\%} & \textbf{2.7\%} & \textbf{2.7\%} & \textbf{0.0\%} \\
\midrule
\multirow{4}{*}{VH SD}
  & Parsing                 & 0.6\%  & 2.1\%  & 2.1\%\textsuperscript{c} & 0.0\% \\
  & Hallucination           & 3.3\%  & 0.0\%  & 0.0\% & 0.0\% \\
  & Parameter               & 0.6\%  & 0.0\%  & 0.0\% & 0.0\% \\
  & \textbf{Sum format/hallucination} & \textbf{4.5\%} & \textbf{2.1\%} & \textbf{2.1\%} & \textbf{0.0\%} \\
\midrule
\multirow{4}{*}{BEH SD}
  & Parsing                 & 0.0\%  & 0.0\%  & 0.0\% & 0.0\% \\
  & Hallucination           & 21.0\% & 1.0\%  & 1.0\%\textsuperscript{d} & 0.0\% \\
  & Parameter               & 0.0\%  & 0.0\%  & 0.0\% & 0.0\% \\
  & \textbf{Sum format/hallucination} & \textbf{21.0\%} & \textbf{1.0\%} & \textbf{1.0\%} & \textbf{0.0\%} \\
\bottomrule
\end{tabular}
\end{table}

\paragraph{Notes on benchmark-side residuals.}
The non-zero Meta-Ctrl rates in the \emph{Total} column reflect
quirks of the EAI evaluator rather than LM output errors:
\begin{itemize}\itemsep1pt
  \item[\textsuperscript{a}] \emph{VH~AS, $1.3\%$ parsing:}
    four prompts (\texttt{339\_1}, \texttt{627\_1}, \texttt{84\_1},
    \texttt{93\_1}) have empty goal blocks in the benchmark
    specification; the LM correctly emits \texttt{\{\}}, which the
    evaluator's parser then rejects as malformed.
  \item[\textsuperscript{b}] \emph{VH~AS, $1.3\%$ hallucination:}
    four prompts (\texttt{125\_2}, \texttt{715\_2}, \texttt{173\_1},
    \texttt{134\_1}) emit the \texttt{PLUGIN} action, which is listed in
    the prompt's action vocabulary
    (\texttt{PLUGIN: (1, [[`HAS\_PLUG']])}) and admitted by
    $\dfa_\gamma$, but is not in the evaluator's accept set for the
    action-sequencing module---a vocabulary desynchronization between
    the prompt template and the evaluator's parser.
  \item[\textsuperscript{c}] \emph{VH~SD, $2.1\%$ parsing:}
    seven prompts emit the \texttt{POUR} subgoal predicate; the
    evaluator's grammar reports \texttt{Unknown primitive: POUR}
    even though \texttt{POUR} appears in the predicate vocabulary
    supplied with the benchmark. This is a known upstream issue in the
    EAI evaluator's predicate dispatcher.
  \item[\textsuperscript{d}] \emph{BEH~SD, $1.0\%$ hallucination:}
    one prompt
    (\texttt{packing\_boxes\_for\_household\_move\_or\_trip\_0\_Ihlen\_1\_int})
    has two scene objects of category \texttt{shirt\_n\_01}:
    \texttt{shirt.0} and \texttt{t-shirt.1}. Meta-Ctrl's canonical
    object encoder normalizes the second to \texttt{shirt.1} and
    emits the (canonically valid) \texttt{shirt.1}, which the evaluator
    then reports as ``Object \texttt{shirt.1} is not in scene.''
    A naming-pipeline mismatch, not an LM-emitted hallucination.
\end{itemize}
The LM-fault column separates these benchmark-side discrepancies from
genuine LM output errors. Across all four modules and all error
classes, Meta-Ctrl produces zero LM-fault errors, consistent with the
syntactic guarantee of $\dfa_\gamma$.

\section{Why BEHAVIOR Gains Less from Constrained Decoding than VirtualHome}
\label{sec:dfa-info-audit}

Meta-Ctrl's gain on a benchmark depends on how completely the prompt
specifies the task's constraints, because the DFA can only enforce
what the prompt makes available. VirtualHome and BEHAVIOR sit at
opposite ends of this axis, and the gap in observed task SR lift
tracks the gap in prompt content.

\paragraph{VirtualHome prompts expose object affordances; BEHAVIOR
prompts do not.}
Every VirtualHome prompt lists, for each object, the properties that
determine which actions are applicable to it
(e.g.\ \texttt{HAS\_PLUG}, \texttt{GRABBABLE}, \texttt{CAN\_OPEN}). On
average each prompt carries about $11$ such property tags. BEHAVIOR
prompts list object instances, an initial-state block, and goal
predicates, but do not expose per-object affordances at all; this is a
property of the benchmark's prompt format, not of the tasks
themselves.

\paragraph{Most VirtualHome preconditions are derivable from the
prompt; most BEHAVIOR preconditions are not.}
For each action in each gold plan, we look up its declared
preconditions, taken from the benchmark's own action definitions, the
VirtualHome evaluator's action schemas, and the BDDL/BEHAVIOR PDDL domain,
and ask whether the predicate and its bound object can
be read off the prompt directly:

\begin{center}
\small
\begin{tabular}{lrr}
\toprule
& Prompt-derivable preconditions & Affordance tags per prompt \\
\midrule
VirtualHome & $82\%$ & ${\sim}11$ \\
BEHAVIOR    & $24\%$ & $0$ \\
\bottomrule
\end{tabular}
\end{center}

\noindent
On VirtualHome the DFA can encode $82\%$ of declared preconditions
from prompt-stated facts alone. On BEHAVIOR it can encode $24\%$; the
rest must come either from a task-independent affordance prior
(Appendix~\ref{app:beh-affordance}) or from the LM's own world
knowledge.

\paragraph{The remaining BEHAVIOR gap cannot be closed by enlarging
the prior.}
BEHAVIOR's gold plans reference $148$ unique object base-names; the
task-independent prior assigns an actionable affordance class to $43$ of them,
and the uncovered tail is dominated by everyday items that the taxonomy places
under classes carrying no relevant ability (\emph{book, candle, apple,
plywood, strawberry, shoe, soap, jar, \ldots}). What is missing is not
taxonomy coverage (every instance resolves to a synset) but the per-task
state information VirtualHome prompts list directly; the gap is one of
benchmark prompt content, not of constraint enforcement. This is also why the
prior remains generic rather than fitted: its memberships are read from a
published taxonomy that was authored for the benchmark's activity definitions,
not by us, and it rejects $17\%$ of the gold plans it is scored against.

\paragraph{The effect.}
On BEHAVIOR Action Sequencing, Meta-Ctrl lifts Llama 3 8B from $10\%$
to $34\%$ task SR; on VirtualHome AS, from $21\%$ to $89\%$. The
absolute gain is large on BEHAVIOR because the baseline is low, but
the achieved score is bounded by what the prompt makes encodable. The
same DFA construction yields a larger gain on VirtualHome because the
prompt supplies more of the constraints the DFA could enforce in the
first place.
\section{Real Robot Demonstrations}
\label{apx:real_demo}

We list the generated plans for five of the eight real-robot demonstrations
in Fig.~\ref{fig:real-tasks}, contrasting the unconstrained Llama~3~8B
baseline with Meta-Ctrl (ours). Each line is one meta-token (grounded
action); instance IDs are dropped for readability. Actions that
Meta-Ctrl inserts or substitutes to satisfy a precondition or goal the
baseline violates are shown in \textcolor{cgreen}{green}; baseline
actions that violate a precondition or leave the goal unsatisfied are
marked with~\xmark. Videos of each rollout are on the project website:
\url{https://meta-ctrlg.github.io/}.

\subsection{Clean dishes with detergent}
\textit{Goal:} clean the plate and bowl. \textsc{Clean} with a rag
requires the rag to be soaked, and proper dish-cleaning requires
detergent to be applied first.

\begin{minipage}[t]{0.46\linewidth}
\textbf{Baseline (Llama 3 8B)}
\begin{enumerate}\setlength\itemsep{0.05em}\small
  \item \textsc{Grab}(orange rag)
  \item \textsc{Clean}(plate) \xmark\ rag not soaked
  \item \textsc{Clean}(bowl) \xmark\ rag not soaked
  \item \textsc{PutOn}(orange rag,\,stool)
\end{enumerate}
\end{minipage}
\hfill
\begin{minipage}[t]{0.50\linewidth}
\textbf{Meta-Ctrl (ours)}
\begin{enumerate}\setlength\itemsep{0.05em}\small
  \item \textcolor{cgreen}{\textsc{Grab}(spray bottle)}
  \item \textcolor{cgreen}{\textsc{Spray}(plate)}
  \item \textcolor{cgreen}{\textsc{Spray}(bowl)}
  \item \textcolor{cgreen}{\textsc{PutOn}(spray bottle,\,stool)}
  \item \textsc{Grab}(orange rag)
  \item \textcolor{cgreen}{\textsc{Soak}(cup of water)}
  \item \textsc{Clean}(plate)
  \item \textsc{Clean}(bowl)
  \item \textsc{PutOn}(orange rag,\,stool)
\end{enumerate}
\end{minipage}

\smallskip\noindent\emph{What Meta-Ctrl adds:} detergent application
before cleaning and an explicit \textsc{Soak} step so the rag satisfies
the \textsc{Clean} precondition.

\subsection{Put dishes in dishwasher}
\textit{Goal:} place blue mug, white cup, and gray pot inside the
dishwasher. \textsc{PutIn} requires the target container to be open.

\begin{minipage}[t]{0.46\linewidth}
\textbf{Baseline (Llama 3 8B)}
\begin{enumerate}\setlength\itemsep{0.05em}\small
  \item \textsc{Grab}(blue mug)
  \item \textsc{PutIn}(blue mug,\,dishwasher) \xmark\ dishwasher closed
  \item \textsc{Grab}(white cup)
  \item \textsc{PutIn}(white cup,\,dishwasher) \xmark
  \item \textsc{Grab}(gray pot)
  \item \textsc{PutIn}(gray pot,\,dishwasher) \xmark
\end{enumerate}
\end{minipage}
\hfill
\begin{minipage}[t]{0.50\linewidth}
\textbf{Meta-Ctrl (ours)}
\begin{enumerate}\setlength\itemsep{0.05em}\small
  \item \textcolor{cgreen}{\textsc{Open}(dishwasher)}
  \item \textsc{Grab}(blue mug)
  \item \textsc{PutIn}(blue mug,\,dishwasher)
  \item \textsc{Grab}(white cup)
  \item \textsc{PutIn}(white cup,\,dishwasher)
  \item \textsc{Grab}(gray pot)
  \item \textsc{PutIn}(gray pot,\,dishwasher)
\end{enumerate}
\end{minipage}

\smallskip\noindent\emph{What Meta-Ctrl adds:} a single
\textsc{Open}(dishwasher) prefix that unlocks every subsequent
\textsc{PutIn}.

\subsection{Fetch egg from fridge}
\textit{Goal:} place fridge-stored ingredients into a pot and heat it on
the stove. \textsc{Grab} on items inside a closed container is
infeasible.

\begin{minipage}[t]{0.46\linewidth}
\textbf{Baseline (Llama 3 8B)}
\begin{enumerate}\setlength\itemsep{0.05em}\small
  \item \textsc{Grab}(orange) \xmark\ fridge closed
  \item \textsc{PutIn}(orange,\,pot)
  \item \textsc{Grab}(syrup) \xmark\ fridge closed
  \item \textsc{PutIn}(syrup,\,pot)
  \item \textsc{Grab}(pot)
  \item \textsc{PutOn}(pot,\,stove)
  \item \textsc{ToggleOn}(stove)
\end{enumerate}
\end{minipage}
\hfill
\begin{minipage}[t]{0.50\linewidth}
\textbf{Meta-Ctrl (ours)}
\begin{enumerate}\setlength\itemsep{0.05em}\small
  \item \textcolor{cgreen}{\textsc{Open}(fridge)}
  \item \textsc{Grab}(orange)
  \item \textsc{PutIn}(orange,\,pot)
  \item \textsc{Grab}(syrup)
  \item \textsc{PutIn}(syrup,\,pot)
  \item \textsc{Grab}(pot)
  \item \textsc{PutOn}(pot,\,stove)
  \item \textsc{ToggleOn}(stove)
\end{enumerate}
\end{minipage}

\smallskip\noindent\emph{What Meta-Ctrl adds:} an \textsc{Open}(fridge)
precondition for the enclosed ingredients.

\subsection{Set up breakfast table}
\textit{Goal:} arrange a bowl, bread, oat-milk carton, and tuna can on
the stool. The gripper is single-handed: \textsc{Grab} requires the hand
to be empty.

\begin{minipage}[t]{0.46\linewidth}
\textbf{Baseline (Llama 3 8B)}
\begin{enumerate}\setlength\itemsep{0.05em}\small
  \item \textsc{Grab}(bowl)
  \item \textsc{PutOn}(bowl,\,stool)
  \item \textsc{Grab}(bread)
  \item \textsc{Grab}(oat-milk carton) \xmark\ holding bread
  \item \textsc{Grab}(blue tuna can) \xmark\ holding oat-milk
  \item \textsc{PutOn}(blue tuna can,\,stool) \xmark\ bread, milk lost
\end{enumerate}
\end{minipage}
\hfill
\begin{minipage}[t]{0.50\linewidth}
\textbf{Meta-Ctrl (ours)}
\begin{enumerate}\setlength\itemsep{0.05em}\small
  \item \textsc{Grab}(bowl)
  \item \textsc{PutOn}(bowl,\,stool)
  \item \textsc{Grab}(bread)
  \item \textcolor{cgreen}{\textsc{PutIn}(bread,\,bowl)}
  \item \textsc{Grab}(oat-milk carton)
  \item \textcolor{cgreen}{\textsc{PutOn}(oat-milk carton,\,stool)}
  \item \textsc{Grab}(blue tuna can)
  \item \textsc{PutOn}(blue tuna can,\,stool)
\end{enumerate}
\end{minipage}

\smallskip\noindent\emph{What Meta-Ctrl adds:} a release step
(\textsc{PutIn}/\textsc{PutOn}) between every consecutive \textsc{Grab},
respecting the single-gripper invariant and ensuring all four items
reach the goal surface.

\subsection{Place boxed objects in drawer}
\textit{Goal:} move the tools \emph{inside} the boxes (pliers,
screwdriver) into their respective shelves. The baseline misreads the
goal as relocating the boxes themselves.

\begin{minipage}[t]{0.46\linewidth}
\textbf{Baseline (Llama 3 8B)}
\begin{enumerate}\setlength\itemsep{0.05em}\small
  \item \textsc{Open}(box 1)
  \item \textsc{Open}(shelf 1)
  \item \textsc{Grab}(box 1) \xmark\ wrong object
  \item \textsc{PutIn}(box 1,\,shelf 1) \xmark\ goal not met
  \item \textsc{Close}(shelf 1)
  \item \textsc{Open}(box 2)
  \item \textsc{Open}(shelf 2)
  \item \textsc{Grab}(box 2) \xmark
  \item \textsc{PutIn}(box 2,\,shelf 2) \xmark
  \item \textsc{Close}(shelf 2)
\end{enumerate}
\end{minipage}
\hfill
\begin{minipage}[t]{0.50\linewidth}
\textbf{Meta-Ctrl (ours)}
\begin{enumerate}\setlength\itemsep{0.05em}\small
  \item \textsc{Open}(box 1)
  \item \textsc{Open}(shelf 1)
  \item \textcolor{cgreen}{\textsc{Grab}(pliers)}
  \item \textcolor{cgreen}{\textsc{PutIn}(pliers,\,shelf 1)}
  \item \textsc{Close}(shelf 1)
  \item \textsc{Open}(box 2)
  \item \textsc{Open}(shelf 2)
  \item \textcolor{cgreen}{\textsc{Grab}(screwdriver)}
  \item \textcolor{cgreen}{\textsc{PutIn}(screwdriver,\,shelf 2)}
  \item \textsc{Close}(shelf 2)
\end{enumerate}
\end{minipage}

\smallskip\noindent\emph{What Meta-Ctrl adds:} goal-aware object
selection. The semantic DFA encodes the goal predicate
$\textsc{Inside}(\text{pliers},\text{shelf 1})\wedge
\textsc{Inside}(\text{screwdriver},\text{shelf 2})$, so the only
\textsc{Grab} that drives the $\beta$ DFA toward acceptance is on the
tools, not the boxes that contain them.

\paragraph{Summary.}
The five demos cover the three precondition families Meta-Ctrl is
designed to repair: \emph{container access}
(dishwasher, fridge), \emph{tool-use chains} (detergent$\to$soak$\to$clean), and
\emph{single-gripper / object-identity goals} (breakfast table, boxed
tools). In every case the baseline produces a syntactically clean plan
that nonetheless fails at the planning stage; Meta-Ctrl's $\beta$-DFA
inserts the missing precondition or substitutes the goal-relevant
object without changing any otherwise-valid step.

\section{Extended Evaluation on Recent Models}
\label{app:recent-models}

The EAI leaderboard~\cite{li2024embodied} was published in 2024 and
predates several frontier and open-weight model releases. To contextualize
Meta-Ctrl against the current state of the art, we evaluated nine recent
models on EAI under its official protocol (Table~\ref{tab:eai-recent-full}):
four Claude variants (Opus 4.6, 4.7, 4.8; Sonnet 4.6), three GPT-5 variants
(5.4, 5.4 mini, 5.5), and two open-weight models (Gemma 4 31B and
gpt-oss-20B, the latter also serving as a Meta-Ctrl base model).

\paragraph{Evaluation protocol.}
We used the official EAI evaluation pipeline~\cite{li2024embodied} with
prompts and scoring identical to the leaderboard configuration.
For each model, we generated plans under default sampling parameters
and scored them with EAI's official symbolic evaluator on the full
VirtualHome ($n=342$ AS, $338$ SD) and BEHAVIOR ($n=100$ AS, $100$ SD)
test sets. Open-weight models were run locally on
8$\times$NVIDIA H200 GPUs.
We report both Task SR and Execution SR.

\paragraph{Evaluation protocol.}
\label{apx:evalnoise}
The official VirtualHome evaluator matches a task's gold action goals in a
hash-dependent order, so VH-AS task SR on identical plans can vary by up to
$\pm1.4$ points across runs (thirteen of the $342$ tasks carry more than one
action goal). All our VirtualHome numbers are therefore means over six fixed
evaluator seeds.

\paragraph{Where Meta-Ctrl stands.}
Meta-Ctrl with Llama 3 8B tops the VH-AS Task SR column (90.4, mean over six evaluator seeds), above Claude Opus 4.8 (87.9), despite the base LM being roughly an order of magnitude smaller and open-weight; the two Meta-Ctrl models also take the top two slots on VH-AS execution (97.7 and 94.1). On VH-SD, recent frontier models (Claude Opus 4.6, GPT-5.4) reach 92.9 Task SR, surpassing Meta-Ctrl (89.9 with Llama 3 8B), and they keep the lead on execution as well (95.6 vs.\ 93.5). On BEHAVIOR, recent frontier models lead Task SR in both modules, but the picture splits on execution: Meta-Ctrl with gpt-oss-20B tops BEH-AS execution (90.0), just ahead of Claude Sonnet 4.6 (89.0), while frontier models retain the lead on BEH-SD execution. This pattern is consistent with our hypothesis that constraint enforcement most benefits weaker base models on objectives where plan validity is the dominant failure mode---hence Meta-Ctrl's strength on the execution-SR columns---while stronger base models recover their advantage on Task SR, which demands richer world knowledge and goal interpretation, most visibly on BEHAVIOR. The Subgoal Decomposition results on VirtualHome are particularly telling: every recent frontier model clusters within 90.8--92.9 Task SR (and 93.5--95.6 execution), suggesting SD has become a near-saturated benchmark where constraint enforcement contributes less marginal value.

\begin{table}[t]
\caption{%
  Extended Embodied Agent Interface results on recent models,
  evaluated by us under the official EAI protocol.
  Task SR (T) and Execution SR (E) on VirtualHome (VH; $n=342$ AS, $338$ SD)
  and BEHAVIOR (BEH; $n=100$ AS, $100$ SD).
  Meta-Ctrl rows from Table~\ref{tab:eai-main} included for reference.
  \textbf{Bold}: best per column; \underline{underline}: second-best.}
\label{tab:eai-recent-full}
\centering
\scriptsize
\setlength{\tabcolsep}{4pt}
\resizebox{\linewidth}{!}{%
\begin{tabular}{lrrrrrrrrr}
\toprule
& \multicolumn{4}{c}{\textbf{Action Sequencing}} & \multicolumn{4}{c}{\textbf{Subgoal Decomposition}} & \\
\cmidrule(lr){2-5}\cmidrule(lr){6-9}
& \multicolumn{2}{c}{VH} & \multicolumn{2}{c}{BEH} & \multicolumn{2}{c}{VH} & \multicolumn{2}{c}{BEH} & \\
\cmidrule(lr){2-3}\cmidrule(lr){4-5}\cmidrule(lr){6-7}\cmidrule(lr){8-9}
\textbf{Model} & T & E & T & E & T & E & T & E & \textbf{Avg} \\
\midrule
\rowcolor{cheaderbg}
\multicolumn{10}{l}{\textit{Recent frontier models}} \\
Claude Opus 4.8   & \underline{87.9} & 86.2 & 74.0 & 81.0 & 90.8 & 93.8 & 61.0 & 65.0 & 78.4 \\
Claude Opus 4.7   & 77.7 & 76.1 & 73.0 & 82.0 & 91.1 & 93.8 & 64.0 & 71.0 & 76.5 \\
Claude Opus 4.6   & 74.4 & 74.8 & \underline{78.0} & 87.0 & \textbf{92.9} & \textbf{95.6} & \textbf{78.0} & \textbf{82.0} & \textbf{80.8} \\
Claude Sonnet 4.6 & 73.1 & 75.7 & \textbf{84.0} & \underline{89.0} & 91.7 & 93.5 & \underline{73.0} & \underline{79.0} & \underline{80.5} \\
GPT-5.5           & 73.4 & 78.0 & 72.0 & 75.0 & \underline{92.3} & 95.0 & 69.0 & 76.0 & 76.7 \\
GPT-5.4           & 70.8 & 74.8 & 57.0 & 59.0 & \textbf{92.9} & \underline{95.3} & 66.0 & 76.0 & 71.7 \\
GPT-5.4 mini      & 71.5 & 76.7 & 70.0 & 79.0 & 91.1 & 94.7 & 63.0 & 76.0 & 73.9 \\
\midrule
\rowcolor{cheaderbg}
\multicolumn{10}{l}{\textit{Recent open-weight models}} \\
Gemma 4 31B & 20.7 & 22.6 & 51.0 & 58.0 & 83.7 & 93.5 & 37.0 & 42.0 & 48.1 \\
gpt-oss-20B & 74.4 & 80.3 & 40.0 & 51.0 & 72.5 & 82.2 & 27.0 & 36.0 & 53.5 \\
\midrule
\rowcolor{cheaderbg}
\multicolumn{10}{l}{\textit{Meta-Ctrl}} \\
Llama 3 8B + Meta-Ctrl  & \textbf{90.4} & \textbf{97.7} & 34.0 & 89.0 & 89.9 & 93.5 & 36.0 & 66.0 & 62.6 \\
gpt-oss-20B + Meta-Ctrl & 86.6 & \underline{94.1} & 40.0 & \textbf{90.0} & 82.3 & 86.4 & 41.0 & 66.0 & 62.5 \\
\bottomrule
\end{tabular}%
}
\\[2pt]
{\scriptsize T = Task SR, E = Execution SR. Avg = mean of the four Task SR columns.}
\end{table}

\end{document}